\documentclass[lettersize,journal]{IEEEtran}
\usepackage{amsmath,amsfonts}
\usepackage{algorithmic}
\usepackage{algorithm}
\usepackage{array}
\usepackage[caption=false,font=normalsize,labelfont=sf,textfont=sf]{subfig}
\usepackage{textcomp}
\usepackage{stfloats}
\usepackage{url}
\usepackage{verbatim}
\usepackage{graphicx}
\usepackage{cite}
\usepackage{graphicx}
\usepackage{amsmath}
\usepackage{amssymb}
\usepackage{booktabs}
\usepackage{url}            
\usepackage{amsfonts}       
\usepackage{nicefrac}       
\usepackage{microtype}      
\usepackage{xcolor}         
\usepackage{graphicx}
\usepackage{array}
\usepackage{multirow}
\usepackage{makecell}
\usepackage{float}

\begin{document}

\title{Attention-based Knowledge Distillation in Multi-attention Tasks: The Impact of a DCT-driven Loss.}

\author{Alejandro L\'{o}pez-Cifuentes, Marcos Escudero-Vi\~nolo, Jes\'{u}s~Besc\'{o}s, Juan C. SanMiguel
        
\thanks{Authors are with the Video Processing and Understanding Lab, Universidad Aut\'{o}noma de Madrid, 28049, Madrid, Spain.}
\thanks{Manuscript received ....}}

\markboth{Journal of \LaTeX\ Class Files,~Vol.~14, No.~8, August~2021}%
{Shell \MakeLowercase{\textit{et al.}}: A Sample Article Using IEEEtran.cls for IEEE Journals}


\maketitle

\begin{abstract}
Knowledge Distillation (KD) is a strategy for the definition of a set of transferability gangways to improve the efficiency of Convolutional Neural Networks. Feature-based Knowledge Distillation is a subfield of KD that relies on intermediate network representations, either unaltered or depth-reduced via maximum activation maps, as the source knowledge. In this paper, we propose and analyse the use of a 2D frequency transform of the activation maps before transferring them. We pose that\textemdash by using global image cues rather than pixel estimates, this strategy enhances knowledge transferability in tasks such as scene recognition, defined by strong spatial and contextual relationships between multiple and varied concepts. To validate the proposed method, an extensive evaluation of the state-of-the-art in scene recognition is presented. Experimental results provide strong evidences that the proposed strategy enables the student network to better focus on the relevant image areas learnt by the teacher network, hence leading to better descriptive features and higher transferred performance than every other state-of-the-art alternative. We publicly release the training and evaluation framework used along this paper at: \url{http://www-vpu.eps.uam.es/publications/DCTBasedKDForSceneRecognition}.
\end{abstract}


\section{Introduction} \label{sec:Introduction}

\IEEEPARstart{D}{eep} Neural Networks, and specifically models based on Convolutional Neural Networks (CNNs), have reached a remarkable success in several computer vision tasks during the last decade \cite{deng2009imagenet, lin2014microsoft, cordts2016cityscapes}. New advances in image databases, CNN architectures and training schemes have pushed forward the state-of-the-art in computer vision. However, the success of deep models, comes usually in hand with the need of huge computational and memory resources to process vast databases for training them \cite{dosovitskiy2020image}. In this vein, there exists a line of research focused on using smaller models that need fewer computational resources for training while obtaining similar results to larger models. Techniques such as quantization \cite{jacob2018quantization}, network pruning \cite{luo2017thinet, zhou2018online, xiao2019autoprune, liu2018rethinking}, Knowledge Distillation \cite{hinton2015distilling, gou2021knowledge} or the design of efficient new architectures \cite{tan2019efficientnet, howard2019searching, cui2019fast} have been of great importance to achieve fast, compact, and easily deploying CNN models.

\subsubsection*{\textbf{Knowledge Distillation}}

Among these, Knowledge Distillation (KD) is of key relevance given its proven effectiveness in different computer vision tasks such as image classification, object detection and semantic segmentation \cite{gou2021knowledge}. KD was originally proposed by Hinton \textit{et al.} \cite{hinton2015distilling} as a strategy to improve the efficiency of CNNs by passing on knowledge from a teacher to a student model. Generally, the student model, usually defined as a smaller network, leverages the knowledge learnt by the teacher model, usually a bigger one, via training supervision. Specifically, in Hinton's KD \cite{hinton2015distilling}, the student model is trained using supervision not only from the ground-truth labels, but also from the teacher predicted logits. Compared to just relying on hard-label annotations, the additional use of teacher's predictions as extra supervision provides an automatic label smoothing regularization \cite{muller2019does,yuan2020revisiting}.

Feature-based Knowledge Distillation expanded the seminal KD scheme by building on the concept of representation learning: CNNs are effective at encoding knowledge at multiple levels of feature representation \cite{bengio2013representation}. The idea was firstly introduced by the FitNets \cite{romero2014fitnets}, which proposed to use the matching of intermediate CNN representations as the source knowledge that is transferred from the teacher to the student. 

\begin{figure*}[t]
    \centering
    \includegraphics[width=0.7\textwidth]{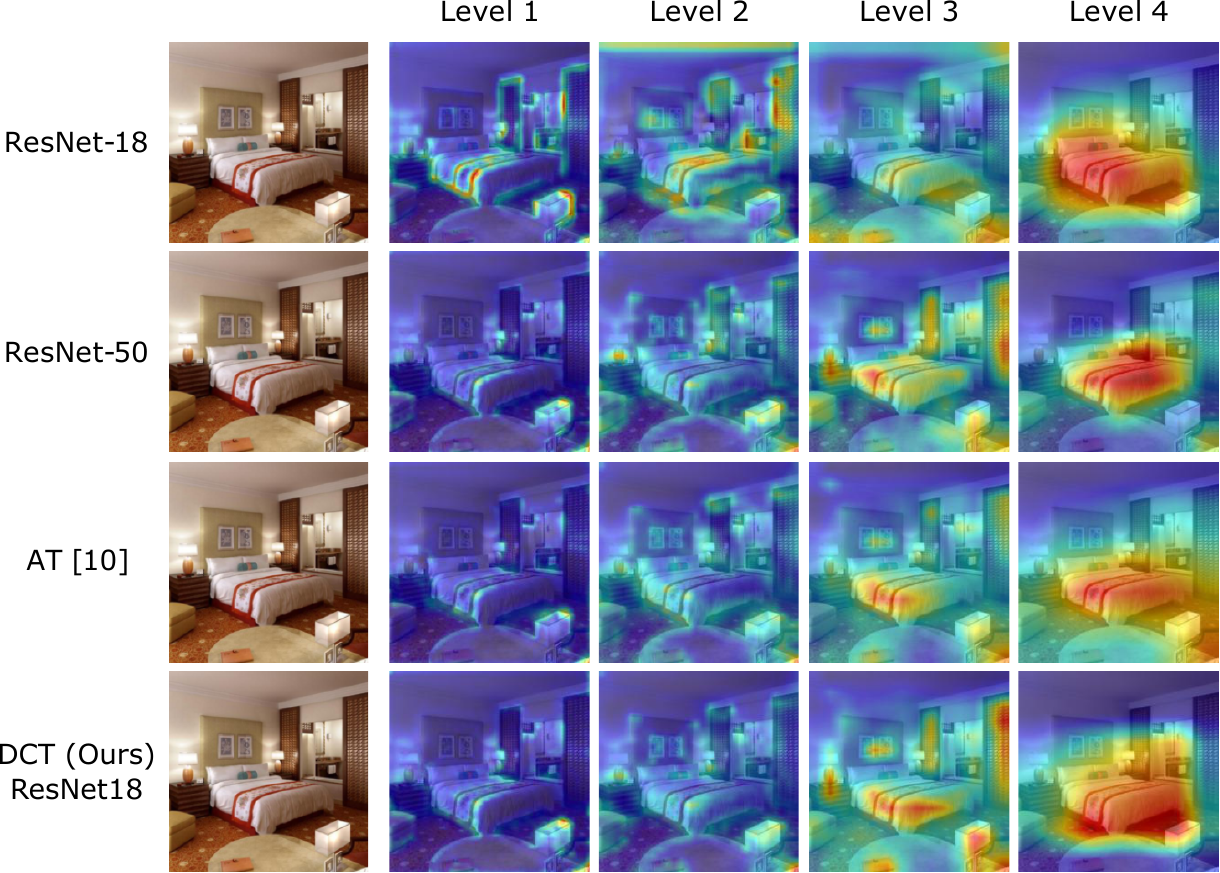}
    \caption{Example of the obtained activation maps, at different levels of depth, for the scene recognition task (the scene class is \textit{hotel room}). Top rows represent activation maps for vanilla ResNet-18 and ResNet-50 CNNs respectively. Bottom row represents the activation maps obtained by the proposed DCT Attention-based KD method when ResNet-50 acts as the teacher network and ResNet-18 acts as the student. AT \cite{komodakis2017paying} activation maps are also included for comparison.}
    \label{fig:ActivationMaps}
\end{figure*}

A specific subgroup of Feature-based KD methods is that of the Attention-based KD ones. This category was pioneered by Komodakis \textit{et al.} \cite{komodakis2017paying}. They proposed to further optimize FitNets by simplifying complete CNN features into attention/activation maps. The matching between the student activation maps and the teacher ones serves as supervision for the KD scheme. The use of activation maps provides several advantages with respect to the direct use of features: first, as matching maps does not depend on channel dimensions, more architectures can be used in the KD process; second, it avoids the problem of semantic mismatching between features when KD is used between two significantly different architectures in terms of depth \cite{chen2020cross}. As depicted in Figure \ref{fig:ActivationMaps}, activation areas, although not being placed in the same image areas, are correlated in terms of the semantic concepts detected even when comparing considerably different models like ResNet-18 and ResNet-50.

Due to its computational simplicity and convenient mathematical properties (differentiable, symmetric and holds the triangle inequality), as already stated by Gou \textit{et al.} \cite{gou2021knowledge}, the convention to compare either two feature tensors or a pair of activation maps is to compute the \(\ell_{2}\) norm of their difference. However, the performance of the \(\ell_{2}\) norm when used to simulate human perception of visual similarities has already been demonstrated to be poor \cite{zhao2015loss}: it might yield, due to its point-wise accumulation of differences, similar results for completely visually different images \cite{wang2009mean}. Furthermore, in the scope of Attention-based KD, another key problem of the \(\ell_{2}\) norm is its tendency towards \textit{desaturation} when is used to guide an optimization process. A visual evidence of this problem is the \textit{sepia} effect in colorization \cite{zhang2016colorful}. We pose that the usage of the pixel-wise \(\ell_{2}\) norm for the comparison of activation maps can be replaced by global image-wise estimates for a better matching and knowledge transferring in Feature-based KD. 

\subsubsection*{\textbf{Contributions}}

In this vein, we propose a novel matching approach based on a 2D discrete linear transform of the activation maps. This novel technique, for which we here leverage the simple yet effective Discrete Cosine Transform (DCT) \cite{oppenheim2001discrete}, is based on the 2D relationships captured by the transformed coefficients, so that the matching is moved from a \textit{pixel-to-pixel} fashion to a correlation in the frequency domain, where each of the coefficients integrates spatial information from the whole image. Figure \ref{fig:ActivationMaps} depicts an example of the obtained activation maps when using the proposed DCT approach to match ResNet-50 ones. Note how the similarity is higher with respect to the ones obtained by AT \cite{komodakis2017paying}, a method based on an \(\ell_{2}\)-driven metric.

In order to verify the effectiveness of the proposed method this paper proposes to use a evaluation of KD in scene recognition, a task defined by strong spatial and contextual relationships among stuff and objects. Scene recognition models are associated to highly variable and sparse attention maps that have been proved to be of crucial relevance for better knowledge modelling and to explain overall performance \cite{lopez2020semantic}. Moreover, we claim that the state-of-the-art in KD is over-fitted to the canonical image classification task (Table \ref{tab:CIFAR100Results}, \cite{chen2021distilling}), where image concepts are represented by a single, usually centered, object (CIFAR and ImageNet datasets). We believe that moving KD research to a more complex task that uses more realistic datasets may be beneficial not only to assess the potential benefits of each KD method in an alternative scenario, but also, to widen the scope of KD research and, in particular, to boost the efficiency of scene recognition models by using models with the same performance but with a significantly lower number of parameters.

In summary, this paper contributes to the KD task by:
\begin{itemize}
    \item Proposing a novel DCT-based metric to compare 2D structures by evaluating their similarity in the DCT domain. We propose to use this technique in an Attention-based KD approach to compare activation maps from intermediate CNN layers more adequately.
    \item Presenting a thorough benchmark of Knowledge Distillation methods on three publicly available scene recognition datasets and reporting strong evidences that the proposed DCT-based metric enables a student network to better focus on the relevant image areas learnt by a teacher model, hence increasing the overall performance for scene recognition.
    \item Publicly releasing the KD framework used to train and evaluate the scene recognition models from the paper. This framework, given its simplicity and modularity, will enable the research community to develop novel KD approaches that can be effortlessly evaluated under the same conditions for scene recognition.
\end{itemize}

\section{Related Work} \label{sec:Related Work}

\subsection{Knowledge-Distillation}
As already introduced, KD is a strategy defining a set of transferability gangways to improve the efficiency of Deep Learning models. A teacher model is used to provide training supervision for a student model, usually a shallower one. Gou \textit{et al.} \cite{gou2021knowledge} proposes to arrange KD into three different groups depending on the \textit{distilled} knowledge: response-based, relation-based and feature-based KD. 

The original KD idea, enclosed in the response-based group, was pioneered by Hinton \textit{et al.} \cite{hinton2015distilling}. They proposed to use teacher outputs in the form of logits to supervise, cooperatively with ground-truth labels, the training of the student network. The training using soft-labels predicted by the teacher provided a strong regularization that benefited the student's performance in the image classification task \cite{muller2019does,yuan2020revisiting}. The seminal KD was improved by changing the way logits were compared. Passalis \textit{et al.} \cite{passalis2020probabilistic} proposed to use a divergence metric (Kullback–Leibler divergence) to match the probability distributions obtained by the teacher and the student. In the same line, Tian \textit{et al.} proposed the use of contrastive learning \cite{tian2019contrastive}, which pushed response-based KD performance even further.

Relation-based KD accounts for transferring the relationships between different activations, neurons or pairs of samples, that are encoded by the teacher model and transferred to the student one. Yim \textit{et al.} \cite{yim2017gift} proposed a Flow of Solution Process (FSP), which is defined by the Gram matrix between two layers. The FSP matrix summarizes the relations between pairs of feature maps. Passalis \textit{et al.} \cite{passalis2020probabilistic} proposed to model abstract feature representations of the data samples by estimating their distribution using a kernel function. Then these estimated distributions were transferred instead of the features, using feature representations of data.

Feature-based KD, as originally proposed by the FitNets transferring scheme \cite{romero2014fitnets}, deals with using the matching of intermediate CNN representations as source knowledge that is transferred from the teacher to the student. Building on top of this idea, a variety of methods have been proposed. Ahn \textit{et al.} \cite{ahn2019variational} formulated feature KD as the maximization of the mutual information between teacher and student features. Guan \textit{et al.} \cite{guan2020differentiable} proposed a student-to-teacher path and a teacher-to-student path to properly obtain feature aggregations. Chen \textit{et al.} \cite{chen2020cross} detected a decrease in performance when distilling knowledge caused by semantic mismatch between certain teacher-student layer pairs, and proposed to use attention mechanisms to automatically weight layers' combinations. Chen \textit{et al.} \cite{chen2021distilling} revealed the importance of connecting features across different levels between teacher and student networks.

Within Feature-based KD methods one can find the attention-based KD ones. Komodakis \textit{et al.} \cite{komodakis2017paying} proposed to simplify the intermediate features to create activation maps that were compared using an \(\ell_{2}\) difference. As already stated in Section \ref{sec:Introduction} and indicated by Gou \textit{et al.} \cite{gou2021knowledge}, it is a convention, not only in attention but also in feature-based KD methods, to build the matching metric based on the \(\ell_{2}\) norm. We argue that this pixel-wise comparison might not be adequate when comparing multi-modal spatial structures such as attention maps.

\subsection{Scene Recognition}
Scene recognition is a hot research topic whose complexity is, according to the reported performances \cite{lopez2020semantic}, one of the highest in image understanding. The complexity of the scene recognition task lies partially on the ambiguity between different scene categories showing similar appearance and objects' distributions: inter-class boundaries can be blurry, as the sets of objects that define a scene might be highly similar to another's. 

Nowadays, top performing strategies are fully based on CNN architectures. Based on context information, Xie \textit{et al.} \cite{xie2017lg} proposed to enhance fine-grained recognition by identifying relevant part candidates based on saliency detection and by constructing a CNN architecture driven by both these local parts and global discrimination. Zhao \textit{et al.} \cite{zhao2018volcano}, similarly, proposed a discriminative discovery network (DisNet) that generates a discriminative map (Dis-Map) for the input image. This map is then used to select scale-aware discriminative locations which are finally forwarded to a multi-scale pipeline for CNN feature extraction.

A specific group of approaches in scene recognition is that trying to model relations between objects information and scenes. Herranz-Perdiguero \textit{et al.} \cite{herranz2018pixels} extended the DeepLab network by introducing SVM classifiers to enhance scene recognition by estimating scene objects and stuff distribution based on semantic segmentation cues. In the same vein, Wang \textit{et al.} \cite{wang2017weakly} defined semantic representations of a given scene by extracting patch-based features from object-based CNNs. The proposed scene recognition method built on these representations\textemdash Vectors of Semantically Aggregated Descriptors (VSAD), ouperformed the state-of-the-art on standard scene recognition benchmarks. VSAD's performance was enhanced by measuring correlations between objects among different scene classes \cite{cheng2018scene}. These correlations were then used to reduce the effect of common objects in scene miss-classification and to enhance the effect of discriminative objects through a Semantic Descriptor with Objectness (SDO). Finally, L\'{o}pez-Cifuentes \textit{et al.} \cite{lopez2020semantic} argued that these methods relied on object information obtained by using patch-based object classification techniques, which entails severe and reactive parametrization (scale, patch-size, stride, overlapping...). To solve this issue they proposed to exploit visual context by using semantic segmentation instead of object information to guide the network's attention. By gating RGB features from information encoded in the semantic representation, their approach reinforced the learning of relevant scene contents and enhanced scene disambiguation by refocusing the receptive fields of the CNN towards the relevant scene contents.

According to the literature, we pose that the differential characteristics of the scene recognition task with respect to classical image classification one might be beneficial to boost and widen the scope of KD techniques. These characteristics include that performance results are not yet saturated, the high ambiguity between different scene categories and that relevant image features are spread out throughout the image instead of being localized in a specific area\textemdash usually the center region of the image.

\section{Attention-based Knowledge Distillation Driven by DCT Coefficients} \label{sec:Method}

\begin{figure*}[t]
    \centering
    \includegraphics[width=0.9\textwidth]{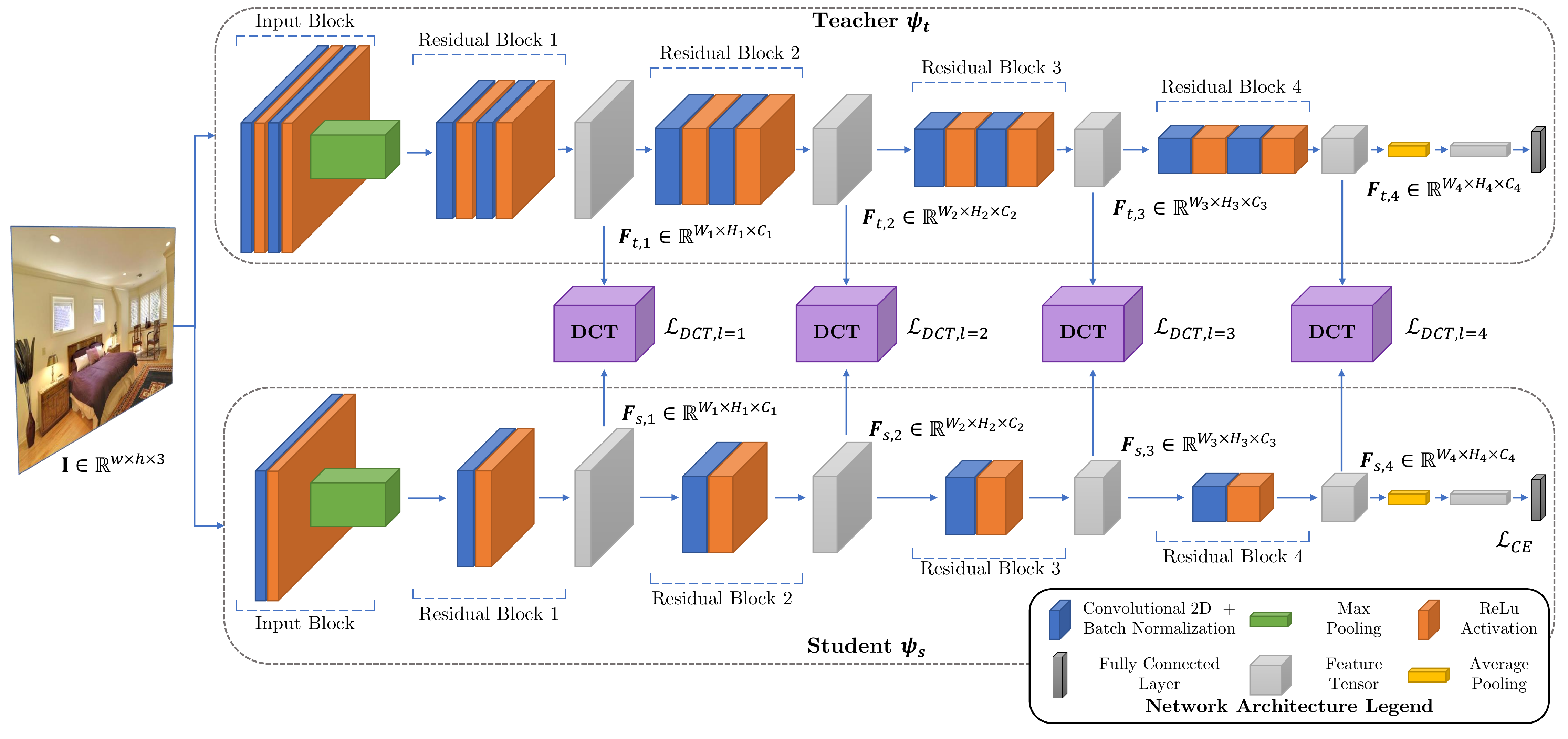}
    \caption{Example of the proposed gangways between two ResNet architectures representing the teacher and the student models. In this case, the intermediate feature representations for the Knowledge Distillation are extracted from the basic Residual Blocks. Besides this example, the proposed method can be applied to the whole set of ResNet, MobileNets, VGGs, ShuffleNets, GoogleNet and DenseNets families.}
    \label{fig:ProposedMethod}
\end{figure*}

Following the organization of KD methods proposed by Gou \textit{\textit{et al.}} \cite{gou2021knowledge}, the following Section is divided into Knowledge (Section \ref{subsec:Knowledge}) and Distillation (Section \ref{subsec:Distillation Scheme}). Figure \ref{fig:ProposedMethod} depicts the proposed DCT gangways in an architecture exemplified with two ResNet branches.

\subsection{Knowledge} \label{subsec:Knowledge}

\textbf{Attention Maps: } We rely on mean feature activation areas \cite{komodakis2017paying}, or attention maps, as the source of knowledge to be transferred from a teacher network to an student network. Given an image \(\textbf{I} \in \mathbb{R}^{3 \times W_I \times H_I}\), a forward pass until a depth \(l\) in a teacher CNN \(\psi_t\) and in a student CNN \(\psi_s\) yields feature tensors \(\psi_t(\mathbf{I},l) = \textbf{F}_{t,l} \in \mathbb{R}^{C_t \times W \times H}\) and \(\psi_s(\mathbf{I},l)=\textbf{F}_{s,l} \in \mathbb{R}^{C_s \times W \times H}\) respectively, with \(W\), \(H\) being the spatial dimensions and \(C_{t}\) and \(C_{s}\) the channel dimensions of the teacher and student features. An activation map for the teacher network \(\textbf{f}_{t,l} \in \mathbb{R}^{W \times H}\) can be obtained from these feature tensors by defining a mapping function \(\mathcal{H}\) that aggregates information from the channel dimensions:

\begin{equation}
    \mathcal{H}: \textbf{F}_{t,l} \in \mathbb{R}^{C_t \times W \times H} \rightarrow \textbf{f}_{t,l} \in \mathbb{R}^{W \times H}.
\end{equation}

The mean squared activations of neurons can be used as an aggregated indicator of the attention of the given CNN with respect to the input image. Accordingly, we define the mapping function \(\mathcal{H}\) as:

\begin{equation}
   \textbf{f}_{t,l} =  \mathcal{H}(\textbf{F}_{t,l}) = \frac{1}{C_t} \sum_{C_{t}} \textbf{F}_{t,l}^2,
\end{equation}

obtaining the feature map \(\textbf{f}_{t,l}\). This activation map is then rescaled to the range \([0, 1]\) by a min-max normalization yielding \(\overline{\textbf{f}}_{t,l}\). This process is similarly applied for the student network to obtain \(\overline{\textbf{f}}_{s,l}\). Figure \ref{fig:ActivationMaps} depicts an example of the normalized activation maps for ResNet-18 and ResNet-50 at different depths. 

\textbf{Comparing Attention Maps via the DCT: } We first propose to apply the DCT \cite{oppenheim2001discrete} to the two activation maps \( \overline{\textbf{f}}_{t,l}\) and \(\overline{\textbf{f}}_{s,l}\) before comparing them. 

For the teacher map, \(\overline{\textbf{f}}_{t,l}\), the DCT yields a set of coefficients \(\mathcal{D}_{t,l} = \lbrace \mathcal{D}(x, y), 0 \leq x, y < W, H \rbrace\), each representing the resemblance or similarity between the whole distribution of \(\overline{\textbf{f}}_{t,l}\) values and a specific 2D pattern represented by the corresponding basis function of the transform. Specifically, in the case of the DCT, these basis functions show increasing variability in the horizontal and vertical dimensions. The DCT is here used over other transformation given its simplicity, its computational efficiency and its differentiability.

Given the lossless nature of the DCT, applying the \(\ell_{2}\) metric to the obtained coefficients of the transformed maps would be equivalent to applying it over the activation maps, as in Komodakis \textit{et al.} \cite{komodakis2017paying}. However, we propose to modify the DCT coefficients in two ways: first, in order to compare the spatial structure of activation maps disregarding the global mean activation we set to zero the first coefficient, the DC coefficient associated to a constant basis function \cite{oppenheim2001discrete}. Then, we rescale the remaining coefficients to the range \([0, 1]\), again using the min-max normalization to obtain \(\overline{\mathcal{D}}_{t,l}\), which permits an scaling of the DCT-term to similar levels of the Cross-Entropy Loss, hence enabling their combination without the need of additional weighting terms. The combination of these three operations (DCT transform, DC coefficient removal and coefficients normalization) in the maps is a simple yet effective change that achieves the comparison to focus on the attention maps distribution rather than on their monomodal maximum. 

After extracting the DCT transform for the student map, the two activation maps are compared using the \(\ell_{2}\) norm between the normalized remaining coefficients by:

\begin{equation} \label{eq:Euclidean Distance DCT}
d_{t,s,l}(\textbf{f}_{t,l}, \textbf{f}_{s,l}) = \sqrt{\sum (\overline{\mathcal{D}}_{t,l} - \overline{\mathcal{D}}_{s,l})^2}.
\end{equation}

With the usage of the \(\ell_{2}\) norm over the DCT coefficients rather than directly on the activation map pixels, we are moving the matching from a pixel-wise computation of differences towards a metric that describes full image differences. In addition, the proposed DCT-based metric focuses on the complete spatial structure while maintaining the mathematical properties of the \(\ell_{2}\) metric: it is a differentiable convex function, it has a distance preserving property under orthogonal transformations and its gradient and Hessian matrix can be easily computed. All of these are desirable and advantageous properties when using this distance in numerical optimization frameworks. 

\subsection{Distillation} \label{subsec:Distillation Scheme}
As stated before, the objective of the proposed distillation scheme is to properly transfer the localization of activation areas for a prediction obtained by the teacher model, $\psi_t$, for a given input \(\textbf{I}\), to the student one, \(\psi_s\). To this aim, we define the KD loss $ \mathcal{L}_{\textrm{\scriptsize DCT}}$ by accumulating the DCT differences along the \(L\) explored gangways:

\begin{equation} \label{eq:Summation L}
    \mathcal{L}_{\textrm{\scriptsize DCT}} = \sum_l^L d_{t,s,l}.
\end{equation}

During training, we refine this loss by only using the teacher maps for correct class predictions. This removes the effect of using distracting maps resulting from teacher's miss-predictions in the knowledge transfer process. In other words, we propose to transfer the knowledge only when the final logit prediction \(\psi_t(\textbf{I})\) is correct. We propose to refine our proposal in Eq. \ref{eq:Summation L} as:

\begin{equation}\label{eq:TeacherPred}
   \mathcal{L}_{\textrm{\scriptsize DCT}}=\left\{
   \begin{array}{ll}
         \sum_l^L d_{t,s,l} & \textrm{ if } \psi_t(\textbf{I}) \textrm{ is correct} \\ [1em]
        0 & \textrm{ else}
    \end{array}\right.
\end{equation}

The overall loss used to train the student CNN \(\psi_s\) is obtained via:
\begin{equation} \label{eq:Final Loss}
    \mathcal{L} = \alpha \mathcal{L}_{\textrm{\scriptsize DCT}} + \beta \mathcal{L}_{\textrm{\scriptsize CE}}, 
\end{equation}

where \(\mathcal{L_{\textrm{\scriptsize CE}}}\) is the regular Cross-Entropy Loss and \(\alpha\) and \(\beta\) are weighting parameters to control the contribution of each term to the final loss.

As usually done with other KD methods \cite{komodakis2017paying, tian2019contrastive, chen2020cross}, the proposed approach can also be combined with the original Response-based KD loss proposed by Hinton \textit{et al.} \cite{hinton2015distilling} by including it in Eq. \ref{eq:Final Loss}:

\begin{equation} \label{eq:Final Loss with KD}
    \mathcal{L} = \alpha \mathcal{L}_{\textrm{\scriptsize DCT}} + \beta \mathcal{L}_{\textrm{\scriptsize CE}} + \delta \mathcal{L}_{\textrm{\scriptsize KD}}, 
\end{equation}

where \(\mathcal{L}_{\textrm{\scriptsize KD}}\) is defined as in Hinton \textit{et al.} \cite{hinton2015distilling} and \(\delta\) weights its contribution to the final loss \(\mathcal{L}\).

\section{Experimental Evaluation} \label{sec:Results}

This Section describes the experiments carried out for validating the proposed approach. First, Section \ref{subsec:Validation on Scene Recognition Benchmarks} delves into the reasons why a new KD benchmark is needed and motivates our choice of the scene recognition task for it. Second, to ease the reproducibility of the method, Section \ref{subsec:Implementation Details} provides a complete review of the implementation details. Section \ref{subsec:Ablation Study} motivates a series of ablation studies for the proposed method. Section \ref{subsec:State-of-the-art} reports state-of-the-art results on the standard CIFAR 100 benchmark and a and thorough state-of-the-art comparison in the scene recognition task. Quantitative and qualitative results for the obtained distilled activation maps are presented in Section \ref{subsec:AnalisysAMs}.

\subsection{Validation on Scene Recognition Benchmarks} \label{subsec:Validation on Scene Recognition Benchmarks}

All feature and attention-based KD methods reviewed in Section \ref{sec:Introduction} and \ref{sec:Related Work} have been mainly evaluated so far using image classification benchmarks on ImageNet \cite{deng2009imagenet}, CIFAR 10/100 \cite{krizhevsky2009learning} and MNIST \cite{deng2012mnist} datasets. We claim that scene recognition is a more suited task to evaluate KD methods for a variety of reasons:

First, reported performances on scene recognition benchmarks \cite{lopez2020semantic, chen2020scene, li2021place} are not saturated. This means that results highly differ between shallow and deep architectures, providing a wider and more representative performance gap to be filled by KD methods than that existing for image classification in standard CIFAR10/100 evaluations. Note how the performance difference between a Teacher and a Vanilla baseline is just a $3\%$ in CIFAR100 (Table \ref{tab:CIFAR100Results}) while that difference grows to a $30\%$ in the ADE20K scene recognition dataset (Table \ref{tab:ADEResults}).

Second, attention is an secondary factor for succeeding in ImageNet-like datasets. Due to the nature of the images, model's attention is usually concentrated around the center of the image \cite{mohsenzadeh2020emergence}. This image-center bias provokes different models focusing on very similar image areas at different depth levels, suggesting that the performance is mainly driven by the representativity and discriminability of the extracted features rather than by the areas of predominant attention. Figure \ref{fig:AMs_CIFAR100} in Section \ref{subsubsec:CIFAR100} provides examples of this observation.

Differently, in scene recognition the gist of a scene is defined by several image features including stuff, objects, textures and spatial relationships between stuff and objects, which are, in turn, spread out throughout the image representing the scene. The areas of attention which different models are primarily focused on have been proved to be critical and to have a strong correlation with performance \cite{lopez2020semantic}. Actually, shallower networks can end up having better performance than deeper networks if their attention is properly guided. In this case, Attention-based KD might be a paramount strategy to build better and simpler models.

Given these reasons, we believe that setting up a KD benchmarking that uses scene recognition rather than classical ImageNet-like image classification is helpful to spread the use of KD to other research scenarios, build a novel state-of-the-art and widen its application to more challenging tasks.

In this section, our approach is evaluated on three well-known and publicly available scene recognition datasets: ADE20K \cite{zhou2017scene}, MIT Indoor 67 \cite{quattoni2009recognizing} and SUN 397 \cite{xiao2010sun}. However, as we understand that our approach should be also compared with respect to KD literature in a standard benchmark, results for CIFAR 100 dataset \cite{krizhevsky2009learning} are also presented in Section \ref{subsubsec:CIFAR100}.

\subsection{Implementation Details} \label{subsec:Implementation Details}
We provide and publicly release a novel training and evaluation KD framework for scene secognition including all the code and methods reported in this paper \footnote{http://www-vpu.eps.uam.es/publications/DCTBasedKDForSceneRecognition}. This framework enables the reproducibility of all the results in the paper and, given its modular design, enables future methods to be easily trained and evaluated under the same conditions as the presented approaches. The following implementation details regarding used architectures, hyper-parameters and evaluation metrics have been used:

\textbf{Architectures:} The proposed method and the state-of-the-art approaches are evaluated using different combinations of Residual Networks \cite{he2016deep} and Mobile Networks \cite{sandler2018mobilenetv2}.

\textbf{Data Normalization and Augmentation:} Each input image is spatially adapted to the network by re-sizing the smaller dimension to \(256\), while the other is resized to mantain the aspect ratio. In terms of data augmentation, we adopt the common data augmentation transformations: random crop to \(224\textrm{x}224\) dimension and random horizontal flipping. We also apply image normalization using ImageNet mean and standard deviation values.

\textbf{Knowledge Distillation Layers}: For the proposed method, we select the intermediate features from ResNets \cite{he2016deep} and MobileNetV2 \cite{sandler2018mobilenetv2} Networks with the following spatial sizes \([H,W]\): \([56,56]\), \([28,28]\), \([14,14]\) and \([7,7]\), analyzing \(L=4\) levels of depth. We assume that both Teacher and Student architectures share the same spatial sizes (in Width and Height, not in Channel dimension) at some points in their architectures. This assumption may preclude the application of the method (to some extent) for pairs of disparate architectures. However, the assumption holds for the most popular architectures (at least those concerning KD and the image classification tasks): the whole set of ResNet, MobileNets, VGGs, ShuffleNets, GoogleNet and DenseNets families. All of these CNN families share the same spatial sizes [H, W] at some points of their architectures. 

\begin{table*}[t]
\renewcommand{\arraystretch}{1.25}
\begin{centering}
\caption{Ablation study regarding different stages of the proposed method. \textit{DCT}: DCT to transform the activation maps. \textit{DC Removal}: suppression of the DC coefficient. \textit{DCT Normalization}: min-max normalization of the DCT coefficients. \textit{Teacher Predictions}: use of teacher predictions to refine the Knowledge Distillation in Eq. \ref{eq:TeacherPred}. Bold values indicate best results.}
\resizebox{\textwidth}{!}{
\begin{tabular}{ccccccccc}
\hline 
DCT & \makecell{DC Removal} & \makecell{DCT Normalization} & \makecell{Teacher Predictions} & \makecell{Hinton's KD \cite{hinton2015distilling}} & Top@1 & Top@5 & MCA & $\Delta$ Top@1 \tabularnewline
\hline 
& &  &  &  & 40.97 & 63.94 & 10.24 & - \tabularnewline
\checkmark{} & &  &  &  & 42.54 & 63.12 & 11.10 & + 3.83 $\%$ \tabularnewline
\checkmark{} &\checkmark{} &  &  &  & 46.51 & 68.92 & 12.45 & + 9.33 $\%$ \tabularnewline
\checkmark{} &\checkmark{} & \checkmark{} &  &  & 46.84 & 67.41 & 12.88 & + 0.70 $\%$ \tabularnewline
\checkmark{} &\checkmark{} & \checkmark{} & \checkmark{} &  & \textbf{47.35} & \textbf{70.40} & \textbf{13.11} & + 1.08 $\%$ \tabularnewline
\checkmark{} &\checkmark{} & \checkmark{} & \checkmark{} & \checkmark{} & \textbf{54.27} & \textbf{76.15} & \textbf{18.05} & + 14.61 $\%$ \tabularnewline
\hline 
\end{tabular}}
\par\end{centering}
\label{tab:Ablation}
\end{table*}

\begin{figure*}[t]
    \centering
    \includegraphics[width=0.7\textwidth,keepaspectratio]{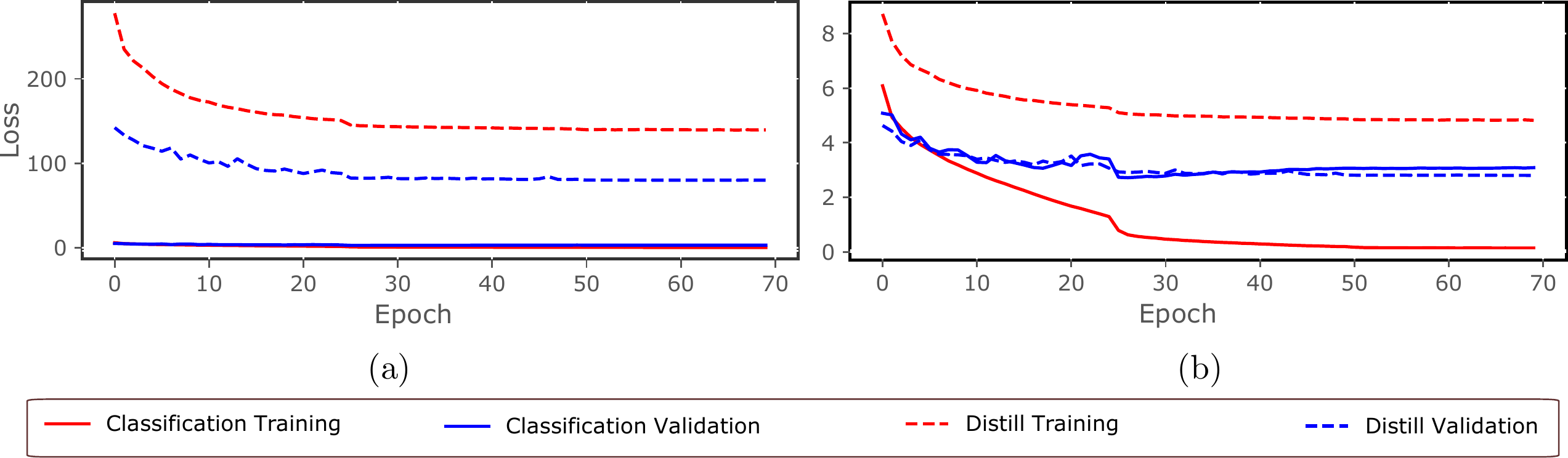}
    \caption{Training and validation losses for ADE20K dataset. Classification curves represent Cross-Entropy loss values. Distill curves represent the proposed DCT-based loss values, either without normalization (a) or using min-max normalization (b).}
    \label{fig:CurvesNormalization}
\end{figure*}

\textbf{Hyper-parameters:} All the reported models have been trained following the same procedure. Stochastic Gradient Descent (SGD) with $0.9$ default momentum and $1^{-4}$ weight decay has been used to minimize the loss function and optimize the student network's trainable parameters. The initial learning rate was set to \(0.1\). All the models have been trained for \(70\) epochs and the learning rate was decayed every \(25\) epochs by a \(0.1\) factor. The batch size was set to \(128\) images. Unless otherwise specified along the Results Section, we set \(\alpha=\beta=1\) in the final loss equation when using the proposed approach. When combining it with Hinton's KD \cite{hinton2015distilling}, we follow the original publication and set \(\beta=0.1\) and \(\delta=1\) while maintaining \(\alpha=1\). All the models, to get rid of potential biases from pretrainings, have been trained from scratch.

All the state-of-the-art reported methods have been trained by us for the scene recognition task using authors' original implementations and implementations from Tian \textit{et al.} \cite{tian2019contrastive}\footnote{\url{https://github.com/HobbitLong/RepDistiller}}. To provide a fair comparison, and in order to adapt them to the scene recognition task, an extensive $\alpha$ grid-search starting from the optimal values reported in the original papers has been performed and presented in Section \ref{subsec:State-of-the-art}. Additionally, for the CIFAR100 experiment in Section \ref{subsubsec:CIFAR100}, optimal hyper-parameter configurations reported in the original papers have been conserved. We refer to each of the individual publications for details.

\textbf{Evaluation Metrics:} Following the common scene recognition procedure \cite{lopez2020semantic}, Top@\(k\) accuracy metric with \(k \in [1, K]\) being \(K\) the total number of Scene classes, has been chosen to evaluate the methods. Specifically, Top@\(\lbrace k=1,5 \rbrace\) accuracy metrics have been chosen. Furthermore, and as the Top@\(k\) accuracy metrics are biased to classes over-represented in the validation set, we also use an additional performance metric, the Mean Class Accuracy (MCA) \cite{lopez2020semantic}. For the CIFAR100 dataset experiment, following \cite{tian2019contrastive} and \cite{chen2021distilling}, regular accuracy is computed.

\textbf{Hardware and Software:} The model design, training and evaluation have been carried out using the PyTorch 1.7.1 Deep Learning framework \cite{paszke2017automatic} running on a PC using a 8 Cores CPU, 50 GB of RAM and a NVIDIA RTX 24GB Graphics Processing Unit.


\subsection{Ablation Studies} \label{subsec:Ablation Study}
The aim of this Section is to gauge the influence of design choices, parameters and computational needs of the method. The performance impact of the different stages of the method are analyzed in Section \ref{subsubsec:Knowledge Distillation Design}, the influence of the $\alpha$ value, that weights the contribution of the proposed DCT-based loss to the global loss function (Eq. \ref{eq:Final Loss}), is measured in Section \ref{subsubsec:Influence of alpha} and the computational overhead introduced by the proposed DCT-based metric is discussed in Section \ref{subsubsec:Computational overhead}.

\subsubsection{Knowledge Distillation Design} \label{subsubsec:Knowledge Distillation Design}

Table \ref{tab:Ablation} quantifies the incremental influence of every step in the proposed approach. For this experiment we use the ADE20K dataset, and ResNet-50 and ResNet-18 for the teacher and student models respectively. Results suggest that even the simplest approach (second row), i.e. when activation maps are distilled from the teacher to the student using the complete non-normalized DCT, outperforms the vanilla baseline (first row). Note that when the DC coefficient is suppressed results are further increased. This suggests that using a metric that captures 2D differences while disregarding the mean intensity value of an activation map helps to increase the performance of the student network. 

Normalization of the DCT coefficients slightly enhances results, but more importantly, scales the DCT loss to be in a similar range than the Cross-Entropy Loss. To further stress the impact of the normalization, Figure \ref{fig:CurvesNormalization} (a) includes loss-evolution graphs for the proposed DCT-based method when DCT coefficients are not normalized, whereas Figure \ref{fig:CurvesNormalization} (b), on the contrary, represents losses when min-max normalization, as described in Section \ref{sec:Method}, is applied prior to the comparison with the $\ell_2$ loss. As it can be observed, the normalization plays a crucial role for scaling the proposed DCT loss. If normalization is not used, the distillation loss term is two orders of magnitude larger than the classification loss term, hence dominating the global loss after their combination. In order to balance the impact of the losses in their combination without normalization, larger $\alpha$ values different than $\alpha=1$ would be required, thereby increasing the complexity of setting adequate hyper-parameters.

Back to Table 1, when Teacher predictions are taken into account and miss-predictions are suppressed from the KD pipeline results are further increased. Finally, the combination of the proposed approach and KD \cite{hinton2015distilling} suggests a high complementarity that can boost results even further.

\subsubsection{Influence of $\alpha$} \label{subsubsec:Influence of alpha}

\begin{figure}[t]
\centering
\centering
\includegraphics[width=0.7\columnwidth]{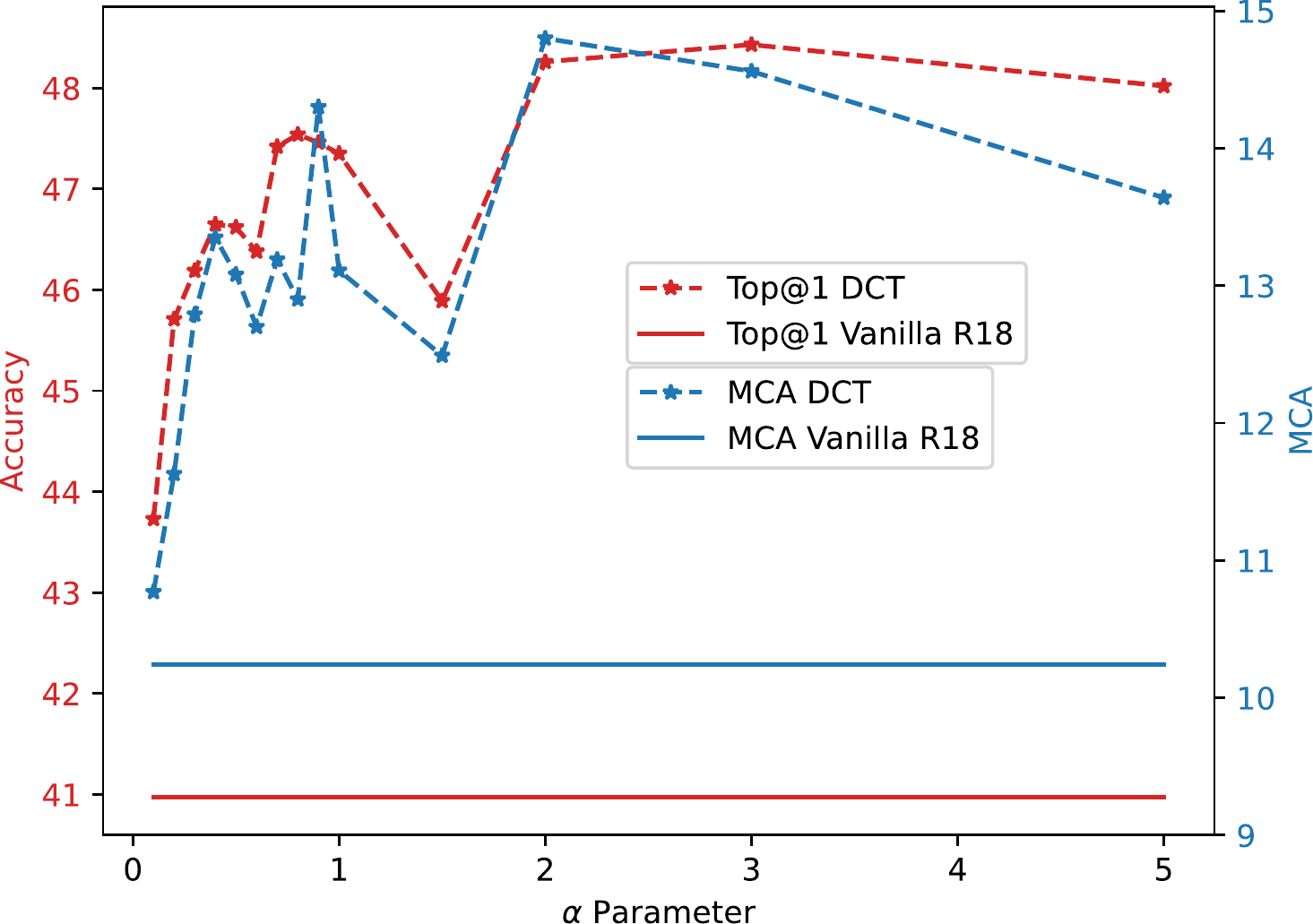}
\caption{Influence of \(\alpha\) in the performance of the model measured over the ADE20K dataset. ResNet-50 acts as the teacher and ResNet-18 as the student.}
\label{fig:AlphaAblation}
\end{figure}

\begin{table}[t]
\renewcommand{\arraystretch}{1.2}
\begin{centering}
\caption{Computational cost comparison measured in extra trainable parameters needed and minutes per training epoch.}
\resizebox{\columnwidth}{!}{
\begin{tabular}{ccc}
\hline 
Method (ResNet-18) & Extra Trainable Parameters & Time per Epoch (Min)\tabularnewline
\hline 
Baseline & - & 0.79\tabularnewline
AT \cite{komodakis2017paying} & 0 M & 1.11\tabularnewline
KD \cite{hinton2015distilling} & 0 M & 1.09\tabularnewline
VID \cite{ahn2019variational} & 12.3 M & 1.53\tabularnewline
Review \cite{chen2021distilling} & 28 M & 1.79\tabularnewline
CKD \cite{chen2020cross} & 634 M & 5.03\tabularnewline
\hline 
DCT (Ours) & 0 M & 1.14\tabularnewline
\hline 
\end{tabular}}
\label{tab:ComputationalCost}
\par\end{centering}
\end{table}

The influence of the \(\alpha\) hyper-parameter (Eq. \ref{eq:Final Loss}) has also been analyzed. Figure \ref{fig:AlphaAblation} shows performance curves (teacher: ResNet-50, student: ResNet-18) obtained with values of \(\alpha\) ranging from \(0.1\) to \(5\) in the ADE20K dataset. For a clearer comparison, performance of the vanilla ResNet-18 is also plotted. It can be observed that our method outperforms vanilla ResNet-18 training for all \(\alpha\) values, suggesting an stable performance for a wide range of \(\alpha\) values. We use \(\alpha=1\) in all the experiments ahead as a trade-off between accuracy and balance of the distillation \(\mathcal{L}_{\textrm{\scriptsize DCT}}\) and the cross-entropy \(\mathcal{L}_{\textrm{\scriptsize CE}}\) terms into the final loss. However, it is important to remark that, differently than reported KD methods that need values of \(\alpha\) ranging usually from \(1\) to \(30000\) (Tables \ref{tab:ADEResults}, \ref{tab:MITResults} and \ref{tab:SUNResults}), the proposed approach is more stable for different $\alpha$ values thanks to the approach described in Section \ref{sec:Method} which facilitates a smooth combination of the \(\mathcal{L}_{\textrm{\scriptsize DCT}}\) and \(\mathcal{L}_{\textrm{\scriptsize CE}}\) losses.

\subsubsection{Computational Overhead} \label{subsubsec:Computational overhead}

Having in mind that computational resources are a key aspect that should be always taken into account, Table \ref{tab:ComputationalCost} presents the overhead derived from including the proposed DCT-based metric with respect to other KD approaches. Results indicate that our approach has a computational time per training epoch similar to that of AT \cite{komodakis2017paying} and KD \cite{hinton2015distilling}. Our implementation leverages the GPU implementation of the Fast Fourier Transform (FFT), which has already been demonstrated to be highly efficient in computational terms. This is also one of the advantages of using the DCT with respect to other alternative transformations. In addition, the proposed method, differently to many others from the state-of-the-art, does not include extra trainable parameters from the student ones, hence not needing extra memory resources.

\subsection{Comparison with the State-of-the-Art} \label{subsec:State-of-the-art}


\subsubsection{\textbf{CIFAR 100 Results}} \label{subsubsec:CIFAR100}

Although one of the aims of our work is to extend and enhance the performance of KD in the scene recognition task, we are aware that an evaluation in the classical KD benchmark on image classification is also needed to help assess our contributions. To this aim, this section presents the performance of the proposed DCT-based approach in the CIFAR-100 dataset. For the sake of consistency, and to provide a fair comparison, we have followed the training and evaluation protocols described in the CRD paper \cite{tian2019contrastive}. In our case, the $\alpha$ parameter from Eq. \ref{eq:Final Loss} has not been modified and remains set to $\alpha=1$. All the performances reported in Table \ref{tab:CIFAR100Results} but those for our method are obtained from already published works \cite{tian2019contrastive, chen2021distilling}.

\begin{table*}[t]
\renewcommand{\arraystretch}{1.25}
\begin{centering}
\caption{CIFAR100 accuracy results with 4 different Teacher-Student combinations. All
the state-of-the-art results are extracted from CRD \cite{tian2019contrastive}
and Review \cite{chen2021distilling} papers. Methods are sorted based
on their average results.}
\resizebox{0.8\textwidth}{!}{
\begin{tabular}{lcccccc}
\hline 
\multirow{2}{*}{Model} & \multirow{2}{*}{Year} & T: ResNet-56 & T: ResNet-110  & T: ResNet-110  & T: ResNet-32x4  & \multirow{2}{*}{\textcolor{blue}{Average}}\tabularnewline
\cline{3-6} 
 &  &  S: ResNet-20 & S: ResNet-20 & S: ResNet-32 & S: ResNet-8x4 & \tabularnewline
\hline 
Teacher & - & 72.34 & 74.31 & 74.31 & 79.42 & \textcolor{blue}{75.09}\tabularnewline
Vanilla & - & 69.04 & 69.06 & 71.14 & 72.50 & \textcolor{blue}{70.43}\tabularnewline
\hline 
RKD \cite{park2019relational} & 2019 & 69.61 & 69.25 & 71.82 & 71.90 & \textcolor{blue}{70.64}\tabularnewline
FitNet \cite{romero2014fitnets} & 2014 & 69.21 & 68.99 & 71.06 & 73.50 & \textcolor{blue}{70.69}\tabularnewline
CC \cite{peng2019correlation} & 2019 & 69.63 & 69.48 & 71.48 & 72.97 & \textcolor{blue}{70.89}\tabularnewline
NST \cite{huang2017like} & 2017 & 69.60 & 69.53 & 71.96 & 73.30 & \textcolor{blue}{71.09}\tabularnewline
FSP \cite{yim2017gift} & 2017 & 69.95 & 70.11 & 71.89 & 72.62 & \textcolor{blue}{71.14}\tabularnewline
FT \cite{kim2018paraphrasing} & 2018 & 69.84 & 70.22 & 72.37 & 72.86 & \textcolor{blue}{71.32}\tabularnewline
SP \cite{tung2019similarity} & 2019 & 69.67 & 70.04 & 72.69 & 72.94 & \textcolor{blue}{71.33}\tabularnewline
VID \cite{ahn2019variational} & 2019 & 70.38 & 70.16 & 72.61 & 73.09 & \textcolor{blue}{71.50}\tabularnewline
AT \cite{komodakis2017paying} & 2017 & 70.55 & 70.22 & 72.31 & 73.44 & \textcolor{blue}{71.63}\tabularnewline
PKT \cite{passalis2020probabilistic} & 2020 & 70.34 & 70.255 & 72.61 & 73.64 & \textcolor{blue}{71.71}\tabularnewline
AB \cite{heo2019knowledge} & 2019 & 69.47 & 69.53 & 70.98 & 73.17 & \textcolor{blue}{71.78}\tabularnewline
KD \cite{hinton2015distilling} & 2015 & 70.66 & 70.67 & 73.08 & 73.33 & \textcolor{blue}{71.93}\tabularnewline
CRD \cite{tian2019contrastive} & 2019 & 71.16 & 71.46 & 73.48 & 75.51 & \textcolor{blue}{72.90}\tabularnewline
\textbf{Review \cite{chen2021distilling}} & \textbf{2021} & \textbf{71.89} & \textbf{71.60} & \textbf{73.89} & \textbf{75.63} & \textbf{\textcolor{blue}{73.25}}\tabularnewline
\hline 
DCT (Ours) & 2022 & 70.45 & 70.10 & 72.42 & 73.52 & \textcolor{blue}{71.55}\tabularnewline
\hline 
\end{tabular}}
\par\end{centering}

\label{tab:CIFAR100Results}
\end{table*}

\begin{table*}[t]
\renewcommand{\arraystretch}{1.2}
\begin{centering}
\caption{ResNet-20 Activation Map's similarity using SSIM with respect to a ResNet-56 model trained
using the CIFAR100 dataset. SSIM values close to 1 indicate identical maps
and values close to 0 indicate no similarity.}
\resizebox{0.8\textwidth}{!}{
\begin{tabular}{lcccccccc}
\hline 
\multirow{2}{*}{Method} & \multicolumn{4}{c}{Training} & \multicolumn{4}{c}{Validation}\tabularnewline
\cline{2-9} 
 & Level 1 & Level 2 & Level 3 & Average & Level 1 & Level 2 & Level 3 & Average\tabularnewline
\hline 
Vanilla ResNet-20 & 0.71 & 0.70 & 0.84 & 0.75 & 0.71 & 0.70 & 0.84 & 0.75\tabularnewline
AT \cite{komodakis2017paying} & 0.92 & 0.92 & \textbf{0.94} & 0.93 & 0.93 & 0.92 & \textbf{0.94} & 0.93\tabularnewline
\textbf{DCT (Ours)} & \textbf{0.97} & \textbf{0.95} & 0.93 & \textbf{0.95} & \textbf{0.97} & \textbf{0.95} & 0.92 & \textbf{0.95}\tabularnewline
\hline 
\end{tabular}}
\par\end{centering}
\label{tab:CIFAR100SSIM}
\end{table*}

Table \ref{tab:CIFAR100Results} presents accuracy results for the state-of-the-art in KD and the proposed approach for several network combinations. To ease the comparison an average column in blue color is also included. These results suggest that: (1) all the reported methods perform similarly: most of them are within the range of $1 \%$ to $3\%$ of accuracy difference; (2) our method achieves results comparable to other state-of-the-art methods even in a single object/concept dataset like CIFAR100.

\begin{figure*}[t]
    \centering
    \includegraphics[width=\textwidth,keepaspectratio]{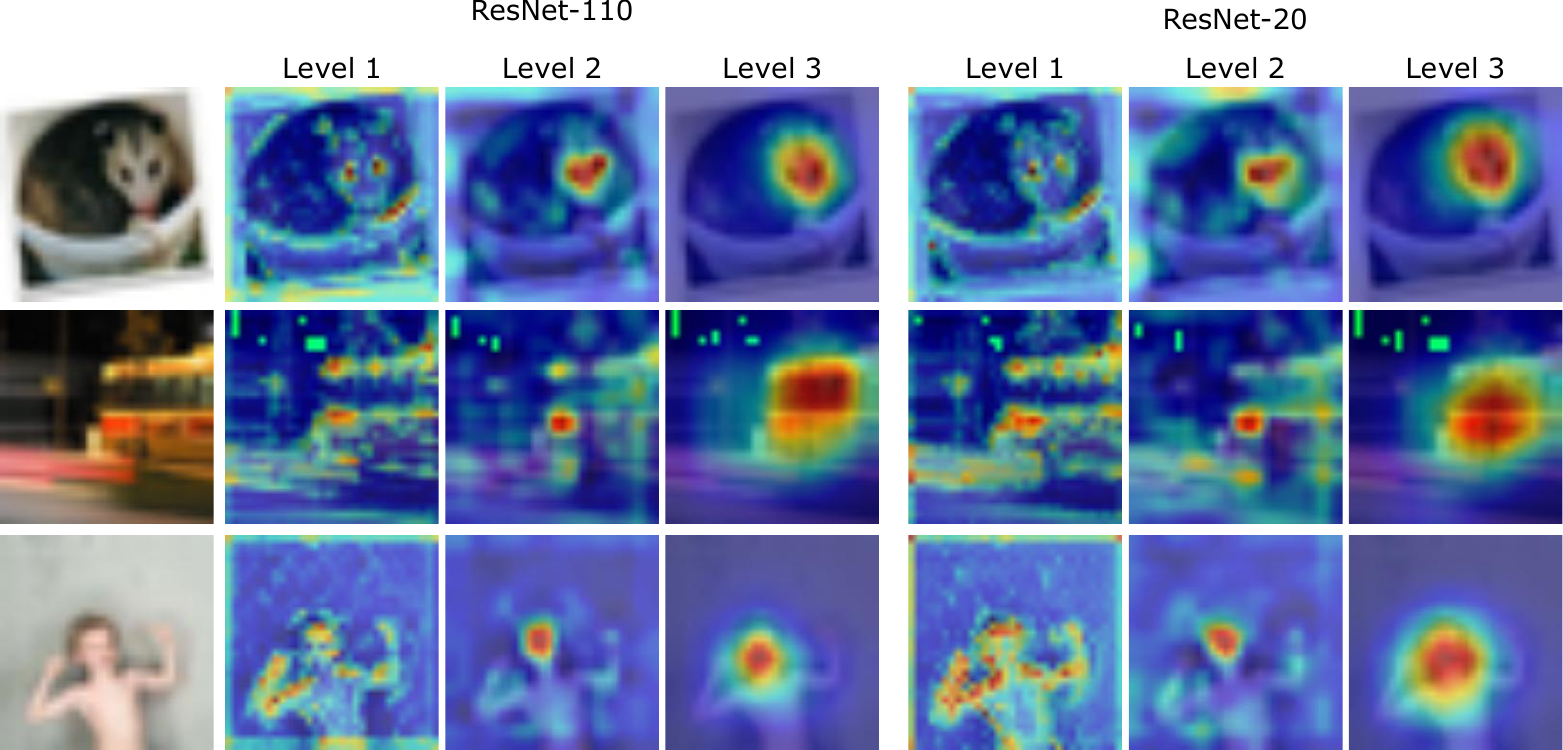}
    \caption{Example of obtained activation maps at three different levels for two different architectures in CIFAR 100 dataset. Note the similarity between activation maps from different architectures and the centered and compact patterns in Level $2$ and Level $3$.}
    \label{fig:AMs_CIFAR100}
\end{figure*}

Our approach is specifically targeted to tasks that benefit from the aggregation of information spatially spread throughout the image, e.g., scene recognition. However, when used for tasks that can be solved just extracting features from a single (usually image-centered) region such as the CIFAR 10/100 image classification benchmark \cite{krizhevsky2009learning}, our proposal is neutral. Contributions from attention-based approaches are hindered due to the similar, centered and compact attention patterns that result from this dataset at all levels of the different CNN vanilla models: as depicted in Figure \ref{fig:AMs_CIFAR100}, highly dissimilar architectures yield similar mono-modal attention maps around the object defining the image class. Note how unlike these attention maps are from the ones depicted in Figure \ref{fig:ActivationMaps}

\begin{figure}[!t]
\centering
\includegraphics[width=0.7\columnwidth]{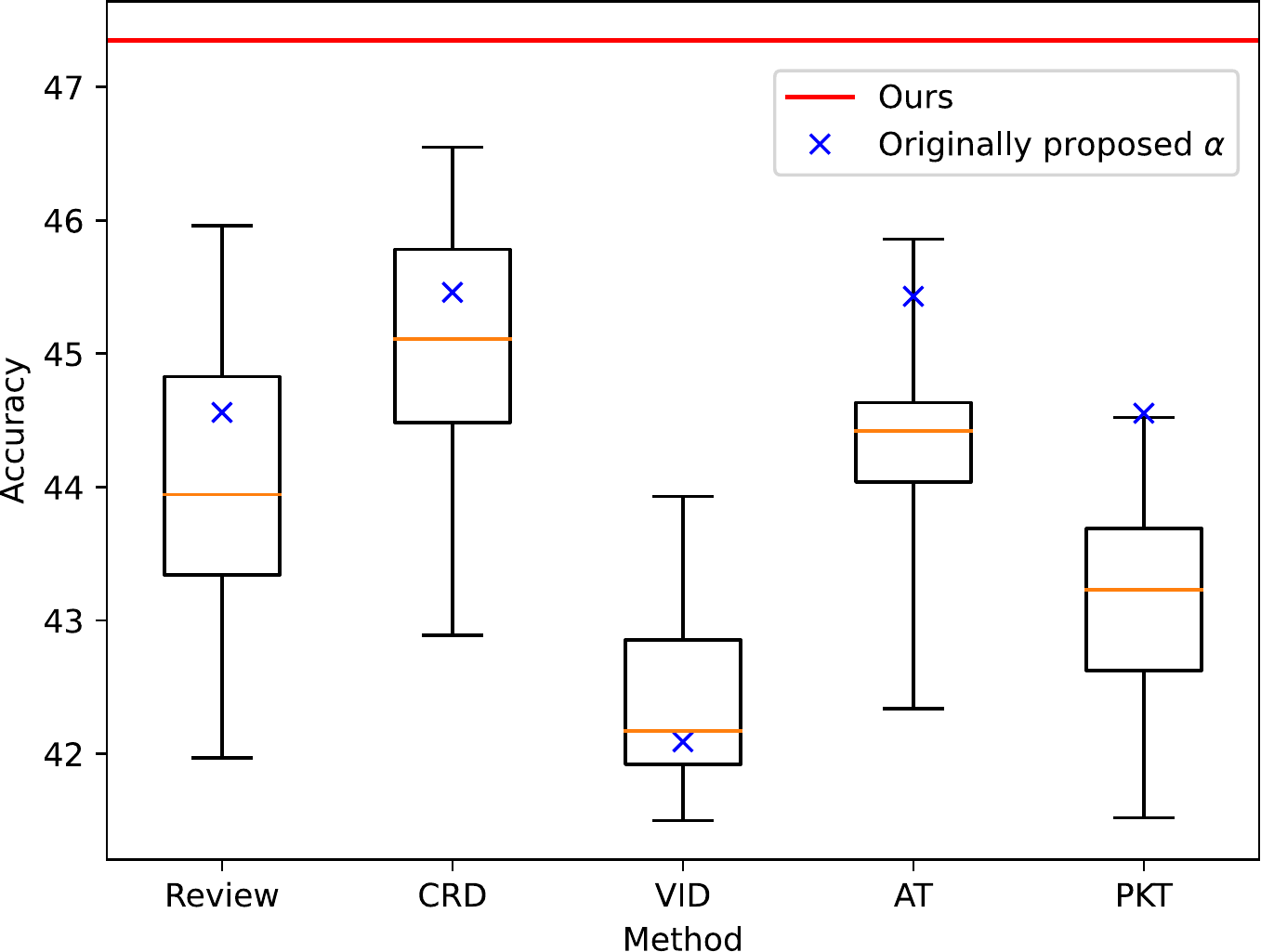}
\caption{Box plot representing state-of-the-art results using $21$ different $\alpha$ values in a range of $\pm100 \%$ from the original value proposed by the corresponding works with a step of $\pm 10 \%$. The study has been performed using ResNet-50 as teacher and ResNet-18 as student in the ADE20K dataset. Red line represents the performance of our approach. Blue crosses represent the performance of each method using the $\alpha$ value reported in the original publications.}
\label{fig:SotaAlphaStudy}
\end{figure}

This attention map bias can be also noticed quantitatively in the experiment reported in Table \ref{tab:CIFAR100SSIM}. Here we quantify the similarity between ResNet-56's (Teacher) and some selected model's activation maps for the whole set of training and validation samples in the CIFAR100 dataset. We use the Structural Similarity Index Measure (SSIM) \cite{wang2004image} to evaluate such similarity, hence avoiding potential biases inherited from the metrics used in the training stage. It can be observed how attention maps for the vanilla ResNet-20 model are, in average, a $75\%$ similar to those of ResNet-56, a model with twice more capacity. It is noteworthy to advance that, when this experiment is carried out for scene recognition (Table \ref{tab:QuantitativeAM}), this average similarity decreases a $36.00\%$ (from 0.75 to 0.48), indicating that the correlation between attention maps is substantially higher for the CIFAR100 than for scene recognition datasets. In other words, activation maps in CIFAR-100 are already matched by most of the methods.

Nevertheless, considering results from Tables \ref{tab:CIFAR100Results} and \ref{tab:CIFAR100SSIM}, one can conclude that the proposed DCT-based loss yields a better matching between Teacher and Student activation maps than a method driven by the $\ell_2$ norm (the AT \cite{komodakis2017paying} method selected for comparison in Table \ref{tab:CIFAR100SSIM}). This supports the motivation of the paper: using a 2D frequency transform of the activation maps before transferring them benefits the comparison of the 2D global information by leveraging the spatial relationships captured by the transformed coefficients.


\begin{figure*}[t]
\centering
\includegraphics[width=\textwidth]{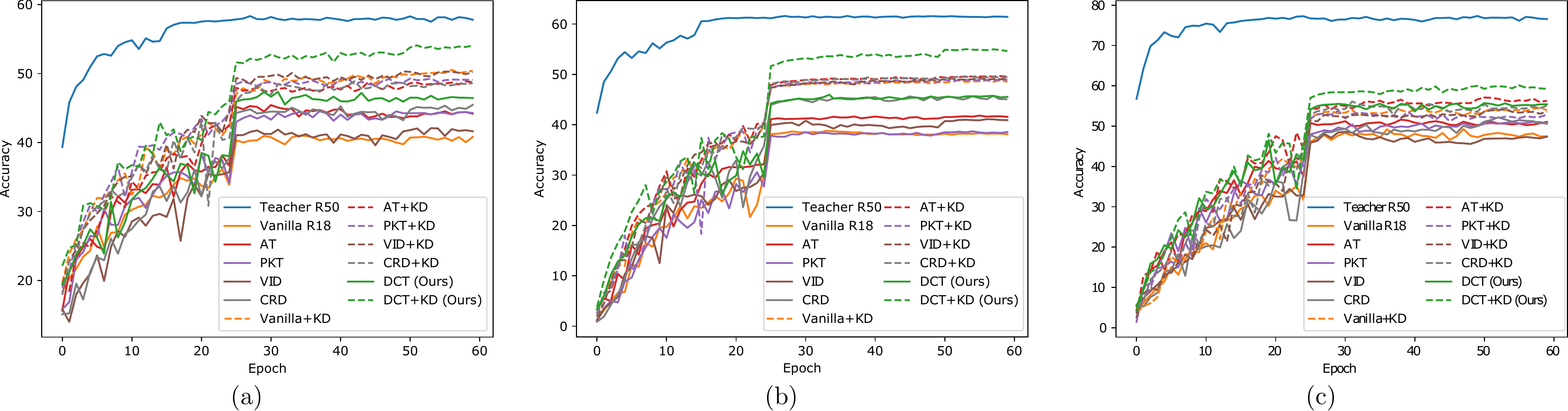}
\caption{Validation set Accuracy (\%) per epoch for the teacher model (ResNet-50), the vanilla network (ResNet-18), state-of-the-art methods, the proposed DCT approach and their combinations with KD \cite{hinton2015distilling} for ADE20K (a), SUN397 (b), and MIT67 (c) datasets.}
\label{fig:ComparativeTrainingCurves}
\end{figure*}

\begin{table*}[!h]
\renewcommand{\arraystretch}{1.25}
\begin{centering}
\caption{Comparison with respect to state-of-the-art methods in the ADE20K dataset with different Teacher (T) - Student (S) combinations. For computational cost comparison the number of additional parameters is indicated. Results are obtained with one run of training. The \(\alpha\) value extracted from Figure \ref{fig:SotaAlphaStudy} and used to train the models is also indicated. Best results in bold.}
\resizebox{\textwidth}{!}{
\begin{tabular}{lcccccccccccc}
\hline 
\multirow{2}{2cm}{Method} & \multirow{2}{0.9cm}{\centering{}Year} & \multirow{2}{1.3cm}{\centering{}Extra Trainable Params} & \multirow{2}{1cm}{\centering{}\(\alpha\)} & \multicolumn{3}{>{\centering}m{3cm}}{T: ResNet-50 (25.6 M)\\S: ResNet-18 (11.7 M)} & \multicolumn{3}{>{\centering}m{3cm}}{T: ResNet-152 (60.3 M)\\ S: ResNet-34 (21.8 M)} & \multicolumn{3}{>{\centering}m{3.2cm}}{T: ResNet-50 (25.6 M)\\S: MobileNet-V2 (3.5 M)}\tabularnewline
\cline{5-13}
 &  & & & Top1 & Top5 & MCA & Top1 & Top5 & MCA & Top1 & Top5 & MCA\tabularnewline
\hline 
Teacher & - & \centering{}- & - & 58.34 & 79.15 & 21.80 & 60.07 & 79.65 & 24.19 & 58.34 & 79.15 & 21.80\tabularnewline
Vanilla & - & \centering{}- & - & 40.97 & 63.94 & 10.24 & 41.63 & 65.15 & 10.03 & 44.29 & 67.69 & 10.44\tabularnewline
\hline 
AT \cite{komodakis2017paying} & 2017 & \centering{}0 M & 1100 & 45.43 & 66.70 & 12.29 & 44.80 & 65.21 & 11.39 & 46.65 & 65.69 & 11.85\tabularnewline
VID \cite{ahn2019variational} & 2019 & \centering{}12.3 M & 1.5 & 43.11 & 65.78 & 10.70 & 41.03 & 62.41 & 9.24 & 43.73 & 66.70 & 10.35\tabularnewline
CRD \cite{tian2019contrastive} & 2019 & \centering{}0.3 M & 1.4 & 45.92 & 67.87 & 11.91 & 43.09 & 66.53 & 10.30 & 45.14 & 69.11 & 10.27\tabularnewline
PKT \cite{passalis2020probabilistic} & 2020 & \centering{}0 M & 30000 & 44.59 & 65.46 & 11.89 & 42.38 & 62.98 & 10.74 & 46.42 & 67.32 & 11.81\tabularnewline
CKD \cite{chen2020cross} & 2021 & \centering{}634 M & 400 & 46.89 & 69.55 & 12.70 & 45.01 & 65.70 & 11.89 & 47.30 & 68.60 & 12.30\tabularnewline
Review \cite{chen2021distilling} & 2021 & \centering{}28 M & 1.8 & 45.88 & 68.20 & 12.71 & 43.03 & 65.34 & 10.84 & 45.30 & 69.74 & 11.48\tabularnewline
\hline 
DCT (Ours) & 2022 & \centering{}0 M & 1 & \textbf{47.35} & \textbf{70.40} & \textbf{13.11} & \textbf{45.63} & \textbf{66.05} & \textbf{12.02} & \textbf{47.39} & \textbf{68.52} & \textbf{12.35}\tabularnewline
\hline 
\hline 
KD \cite{hinton2015distilling} & 2015 & \centering{}0 M & 0.8 & 50.54 & 73.49 & 15.39 & 48.91 & 73.37 & 14.51 & 48.37 & 71.47 & 12.55\tabularnewline
AT \cite{komodakis2017paying} + KD & 2017 & \centering{}0 M & 1100 & 48.87 & 73.01 & 13.29 & 49.35 & 72.09 & 14.16 & 47.67 & 72.97 & 12.93\tabularnewline
VID \cite{ahn2019variational} + KD & 2019 & \centering{}12.3 M & 1.5 & 49.69 & 72.36 & 19.89 & 49.34 & 71.57 & 14.19 & 48.14 & 71.88 & 12.90\tabularnewline
CRD \cite{tian2019contrastive} + KD & 2019 & \centering{}0.3 M & 1.4 & 48.78 & 73.76 & 12.31 & 48.16 & 72.15 & 15.36 & 47.88 & 71.97 & 11.36\tabularnewline
PKT \cite{passalis2020probabilistic} + KD & 2020 & \centering{}0 M & 30000 & 49.31 & 73.41 & 14.48 & 49.70 & 73.33 & 14.64 & 49.43 & 72.76 & 13.59\tabularnewline
CKD \cite{chen2020cross} + KD & 2021 & \centering{}634 M & 400 & 52.10 & 76.90 & 15.54 & \textbf{53.54} & \textbf{75.20} & \textbf{17.98} & 49.15 & 70.25 & 13.32\tabularnewline
Review \cite{chen2021distilling} + KD & 2021 & \centering{}28 M & 1.8 & 50.63 & 73.73 & 14.86 & 49.59 & 72.56 & 14.99 & 48.32 & 71.84 & 12.12\tabularnewline
\hline 
DCT (Ours) + KD & 2022 & \centering{}0 M & 1 & \textbf{54.25} & \textbf{76.15} & \textbf{18.05} & 52.68 & 74.60 & 17.07 & \textbf{50.75} & \textbf{72.53} & \textbf{14.05}\tabularnewline
\hline
\end{tabular}}
\label{tab:ADEResults}
\par\end{centering}
\end{table*}

\subsubsection{\textbf{Scene Recognition Results}}
This Section presents a state-of-the-art benchmark for KD methods. Following common evaluations \cite{tian2019contrastive, chen2020cross, chen2021distilling} we have selected top performing KD methods: KD \cite{hinton2015distilling}, AT \cite{komodakis2017paying}, PKT \cite{passalis2020probabilistic}, VID \cite{ahn2019variational}, CRD \cite{tian2019contrastive}, CKD \cite{chen2020cross} and Review \cite{chen2021distilling}. Obtained results for ADE20K, SUN397 and MIT67 datasets are presented in Tables \ref{tab:ADEResults}, \ref{tab:SUNResults} and \ref{tab:MITResults} respectively. Performance metrics are included for three different pairs of teacher/student models: two sharing the same architecture, ResNet-50/ResNet-18 and ResNet-152/ResNet-34, and one with different backbones, ResNet-50/MobileNetV2. In addition, the combination of all these models with Hinton's KD \cite{hinton2015distilling} is also reported.

First, to provide a fair comparison, Figure \ref{fig:SotaAlphaStudy} compiles the performance ranges of an extensive search of the optimal $\alpha$ value for each of the compared methods for the scene recognition task. The search has been carried out modifying the $\alpha$ values reported in the original publications (which we understand optimal for the image classification task) in a range between $\pm100 \%$ with a step of $\pm 10 \%$. The search has been performed using ResNet-50 as teacher and ResNet-18 as student in the ADE20K dataset. To ease the comparison, the performance obtained by the original $\alpha$ value and the proposed method is also included. The models trained using $\alpha$ values resulting in the best performance for each method have been used to obtain the results from Tables \ref{tab:ADEResults}, \ref{tab:SUNResults} and \ref{tab:MITResults}.

Average results from Tables \ref{tab:ADEResults}, \ref{tab:SUNResults} and \ref{tab:MITResults} indicate that the proposed approach outperforms both the vanilla training of the student and all the reported KD methods. The training loss curves for the validation sets depicted in Figures \ref{fig:ComparativeTrainingCurves} (a), \ref{fig:ComparativeTrainingCurves}(b) and \ref{fig:ComparativeTrainingCurves} (c) support this assumption providing a graphical comparison between all the reported methods for ADE20K, SUN397 and MIT67 datasets respectively.

\begin{table*}[!h]
\renewcommand{\arraystretch}{1.25}
\begin{centering}
\caption{Comparison with respect to state-of-the-art methods in the SUN397 dataset with different Teacher (T) - Student (S) combinations.}
\resizebox{\textwidth}{!}{
\begin{tabular}{lcccccccccccc}
\hline 
\multirow{2}{2cm}{Method} & \multirow{2}{0.9cm}{\centering{}Year} & \multirow{2}{1.3cm}{\centering{}Extra Trainable Params} & \multirow{2}{1cm}{\centering{}\(\alpha\)} & \multicolumn{3}{>{\centering}m{3cm}}{T: ResNet-50 (25.6 M)\\S: ResNet-18 (11.7 M)} & \multicolumn{3}{>{\centering}m{3cm}}{T: ResNet-152 (60.3 M)\\ S: ResNet-34 (21.8 M)} & \multicolumn{3}{>{\centering}m{3.2cm}}{T: ResNet-50 (25.6 M)\\S: MobileNet-V2 (3.5 M)}\tabularnewline
\cline{5-13}
 &  & & & Top1 & Top5 & MCA & Top1 & Top5 & MCA & Top1 & Top5 & MCA\tabularnewline
\hline 
Teacher & - & \centering{}- & - & 61.69 & 87.50 & 61.74 & 62.56 & 87.53 & 62.63 & 61.69 & 87.50 & 61.74\tabularnewline
Vanilla & - & \centering{}- & - & 38.77 & 67.05 & 38.83 & 39.66 & 69.36 & 40.10 & 41.18 & 70.58 & 41.23\tabularnewline
\hline 
AT \cite{komodakis2017paying} & 2017 & \centering{}0 M & 1100 & 41.52 & 69.87 & 41.58 & 40.75 & 69.53 & 40.81 & 38.84 & 68.08 & 38.91\tabularnewline
VID \cite{ahn2019variational} & 2019 & \centering{}12.3 M & 1.5 & 41.16 & 69.15 & 41.21 & 39.02 & 67.77 & 39.05 & 40.59 & 69.79 & 40.64\tabularnewline
CRD \cite{tian2019contrastive} & 2019 & \centering{}0.3 M & 1.4 & 43.89 & 73.55 & 43.95 & 42.13 & 71.51 & 42.14 & 42.69 & 72.98 & 42.73\tabularnewline
PKT \cite{passalis2020probabilistic} & 2020 & \centering{}0 M & 30000 & 38.70 & 67.34 & 38.72 & 37.70 & 66.06 & 37.72 & 40.17 & 68.89 & 40.2\tabularnewline
Review \cite{chen2021distilling} & 2021 & \centering{}28 M & 1.8 & 43.26 & 72.77 & 43.29 & 42.69 & 70.92 & 42.73 & 42.68 & 71.72 & 42.74\tabularnewline
\hline 
DCT (Ours) & 2022 & \centering{}0 M & 1 & \textbf{45.75} & \textbf{74.59} & \textbf{45.80} & \textbf{43.50} & \textbf{72.33} & \textbf{43.54} & \textbf{43.16} & \textbf{70.59} & \textbf{43.19}\tabularnewline
\hline 
\hline 
KD \cite{hinton2015distilling} & 2015 & \centering{}0 M & 0.8 & 48.83 & 77.66 & 48.90 & 48.26 & 76.79 & 48.30 & 47.31 & 77.80 & 47.38\tabularnewline
AT \cite{komodakis2017paying} + KD & 2017 & \centering{}0 M & 1100 & 49.44 & 78.06 & 49.52 & 47.05 & 75.39 & 49.10 & 46.60 & 76.42 & 46.08\tabularnewline
VID \cite{ahn2019variational} + KD & 2019 & \centering{}12.3 M & 1.5 & 49.26 & 78.16 & 49.32 & 47.08 & 75.95 & 47.12 & 46.64 & 76.87 & 46.71\tabularnewline
CRD \cite{tian2019contrastive} + KD & 2019 & \centering{}0.3 M & 1.4 & 49.79 & 78.69 & 49.82 & 48.39 & 77.00 & 48.44 & 46.77 & 77.30 & 46.82\tabularnewline
PKT \cite{passalis2020probabilistic} + KD & 2020  & \centering{}0 M & 30000 & 49.13 & 78.16 & 49.16 & 48.08 & 76.75 & 48.15 & 47.54 & 77.51 & 47.56\tabularnewline
Review \cite{chen2021distilling} + KD & 2021 & \centering{}28 M & 1.8 & 49.90 & 78.71 & 49.96 & 47.05 & 76.30 & 47.07 & 47.05 & 77.44 & 47.10\tabularnewline
\hline 
DCT (Ours) + KD & 2022 & \centering{}0 M & 1 & \textbf{55.15} & \textbf{83.20} & \textbf{55.19} & \textbf{50.51} & \textbf{79.25} & \textbf{50.55} & \textbf{49.25} & \textbf{79.35} & \textbf{49.30}\tabularnewline
\hline 
\end{tabular}}
\label{tab:SUNResults}
\par\end{centering}
\end{table*}

\begin{table*}[!h]
\renewcommand{\arraystretch}{1.25}
\begin{centering}
\caption{Comparison with respect to state-of-the-art methods in the MIT67 dataset with different Teacher (T) - Student (S) combinations.}
\resizebox{\textwidth}{!}{
\begin{tabular}{lcccccccccccc}
\hline 
\multirow{2}{2cm}{Method} & \multirow{2}{0.9cm}{\centering{}Year} & \multirow{2}{1.3cm}{\centering{}Extra Trainable Params} & \multirow{2}{1cm}{\centering{}\(\alpha\)} & \multicolumn{3}{>{\centering}m{3cm}}{T: ResNet-50 (25.6 M)\\S: ResNet-18 (11.7 M)} & \multicolumn{3}{>{\centering}m{3cm}}{T: ResNet-152 (60.3 M)\\ S: ResNet-34 (21.8 M)} & \multicolumn{3}{>{\centering}m{3.2cm}}{T: ResNet-50 (25.6 M)\\S: MobileNet-V2 (3.5 M)}\tabularnewline
\cline{5-13}
 &  & & & Top1 & Top5 & MCA & Top1 & Top5 & MCA & Top1 & Top5 & MCA\tabularnewline
\hline 
Teacher & - & \centering{}- & - & 77.32 & 95.20 & 79.00 & 78.11 & 95.02 & 78.91 & 77.32 & 95.20 & 79.00\tabularnewline
Vanilla & - & \centering{}- & - & 49.26 & 77.02 & 46.87 & 38.84 & 67.52 & 38.88 & 49.06 & 79.08 & 48.66\tabularnewline
\hline 
AT \cite{komodakis2017paying} & 2017 & \centering{}0 M & 1100 & 50.41 & 79.30 & 50.42 & 49.66 & 76.84 & 49.03 & 45.13 & 75.51 & 44.32\tabularnewline
VID \cite{ahn2019variational} & 2019 & \centering{}12.3 M & 1.5 & 48.21 & 76.71 & 47.60 & 44.22 & 72.77 & 43.23 & 47.76 & 75.96 & 47.14\tabularnewline
CRD \cite{tian2019contrastive} & 2019 & \centering{}0.3 M & 1.4 & 51.45 & 78.56 & 51.14 & 41.95 & 72.95 & 41.87 & 50.10 & 77.22 & 47.20\tabularnewline
PKT \cite{passalis2020probabilistic} & 2020 & \centering{}0 M & 30000 & 51.03 & 79.15 & 49.56 & 46.32 & 74.34 & 45.55 & 50.23 & 78.80 & 47.92\tabularnewline
Review \cite{chen2021distilling} & 2021 & \centering{}28 M & 1.8 & 51.73 & 80.78 & 51.18 & 44.43 & 75.36 & 44.09 & 50.25.48 & 78.60 & 49.43\tabularnewline
\hline 
DCT (Ours) & 2022 & \centering{}0 M & 1 & \textbf{56.32} & \textbf{84.90} & \textbf{55.39} & \textbf{52.14} & \textbf{80.98} & \textbf{50.98} & \textbf{50.42} & \textbf{78.68} & \textbf{48.38}\tabularnewline
\hline 
\hline 
KD \cite{hinton2015distilling} & 2015 & \centering{}0 M & 0.8 & 54.87 & 83.42 & 54.91 & 51.55 & 79.61 & 51.24 & 56.14 & 82.51 & 56.04\tabularnewline
AT \cite{komodakis2017paying} + KD & 2017 & \centering{}0 M & 1100 & 58.41 & 83.78 & 57.81 & 52.30 & 80.10 & 52.48 & 52.17 & 80.53 & 51.34\tabularnewline
VID \cite{ahn2019variational} + KD & 2019 & \centering{}12.3 M & 1.5 & 54.20 & 81.51 & 54.54 & 51.79 & 80.23 & 51.88 & 55.75 & 81.94 & 55.60\tabularnewline
CRD \cite{tian2019contrastive} + KD & 2019 & \centering{}0.3 M & 1.4 & 55.23 & 83.83 & 54.83 & 50.54 & 79.92 & 50.53 & 55.16 & 81.78 & 54.79\tabularnewline
PKT \cite{passalis2020probabilistic} + KD & 2020 & \centering{}0 M & 30000 & 53.83 & 80.83 & 53.77 & 50.52 & 79.37 & 50.71 & 53.05 & 81.87 & 52.90\tabularnewline
Review \cite{chen2021distilling} + KD & 2021 & \centering{}28 M & 1.8 & 56.48 & 81.89 & 57.17 & 51.42 & 78.96 & 51.05 & 56.99 & 81.59 & 56.98\tabularnewline
\hline 
DCT (Ours) + KD & 2022 & \centering{}0 M & 1 & \textbf{60.11} & \textbf{86.88} & \textbf{60.53} & \textbf{55.18} & \textbf{81.64} & \textbf{55.62} & \textbf{57.35} & \textbf{84.79} & \textbf{56.89}\tabularnewline
\hline 
\end{tabular}}
\label{tab:MITResults}
\par\end{centering}
\end{table*}

Results from the proposed method compared with respect to the rest of the approaches reinforce the hypothesis  that properly learnt CNN attention is crucial for scene recognition. Results from smaller networks can be boosted if their attention is properly guided towards representative image areas, which are better obtained by deeper and more complex architectures. The increase in performance of the method with respect to AT \cite{komodakis2017paying} suggests that, even though adopting similar knowledge sources, the proposed loss is able to consistently achieve better results by better quantifying the differences between attention maps. 

CKD \cite{chen2020cross} outperforms our method in an specific combination of Table \ref{tab:ADEResults} (T: ResNet-152 and S: ResNet-34 + KD) for the ADE20K dataset, being behind us in the other two combinations evaluated. Nevertheless, the number of extra trainable parameters required by CKD grows with the resolution of the images: whereas CKD is reasonable for datasets composed of low-resolution images (CIFAR 10/100 datasets), here the number of parameters is $30$ times larger than the teacher from where the knowledge is transferred. Given this amount of extra trainable parameters, it may be worthier to train a vanilla model with that capacity. Therefore, we do not include the evaluation for CKD in the SUN397 and MIT67 datasets.

Results from Tables \ref{tab:ADEResults}, \ref{tab:SUNResults} and \ref{tab:MITResults} also indicate that when dealing with scene recognition datasets a proper selection of the architectures to be used in KD is important. Note how using a deeper architecture like ResNet-152 might not be as beneficial as using ResNet-50, maybe due to overfitting, or how extremely efficient models like MobileNet-V2 can get similar results as ResNet-18 or ResNet-34.

When the proposed method is combined with KD \cite{hinton2015distilling}, results show an increase in performance with respect to the rest of the methods, which evidences that the proposed DCT-based method can be properly combined with KD, benefiting from the extra regularization that seminal KD provides at the response level.


\begin{table}[t]
\scriptsize
\renewcommand{\arraystretch}{1.2}
\begin{centering}
\caption{Error Rates when transfering learning from ImageNet to MIT67 scene recognition Dataset. All results except DCT (Ours) are extracted from Zagoruyko \textit{et al.} \cite{komodakis2017paying}.}
\resizebox{0.7\columnwidth}{!}{
\begin{tabular}{llc}
\hline 
Method & Backbone & Error Rate\tabularnewline
\hline 
Teacher & ResNet-34 & 26.0\tabularnewline
Student & ResNet-18 & 28.2\tabularnewline
AT \cite{komodakis2017paying} & ResNet-18 & 27.1\tabularnewline
KD \cite{hinton2015distilling} & ResNet-18 & 28.1\tabularnewline
DCT (Ours) & ResNet-18 & \textbf{26.35}\tabularnewline
\hline 
\end{tabular}}
\par\end{centering}
\label{tab:TransferLearning}
\end{table}

\subsubsection{Transfer Learning Results}
Table \ref{tab:TransferLearning} presents a Transfer Learning experiment for scene recognition. We have followed the same training and evaluation protocol for the AT method as that proposed by Zagoruyko \textit{et al.} \cite{komodakis2017paying}. The aim of the experiment is to illustrate that our method also works when transferring attention in a Transfer Learning scenario, i.e., fine tuning a model to the MIT67 dataset from a model with ImageNet pre-trained weights. Results indicate that the proposed approach helps the transfer learning process by decreasing the error rate a \(6.56\%\) and a \(2.76\%\) with respect to the student and AT-transferred model, respectively.


\subsection{Analysis of Activation Maps}\label{subsec:AnalisysAMs}

\begin{figure*}[!t]
    \centering
    \includegraphics[width=0.8\textwidth, height=0.8\textheight, keepaspectratio]{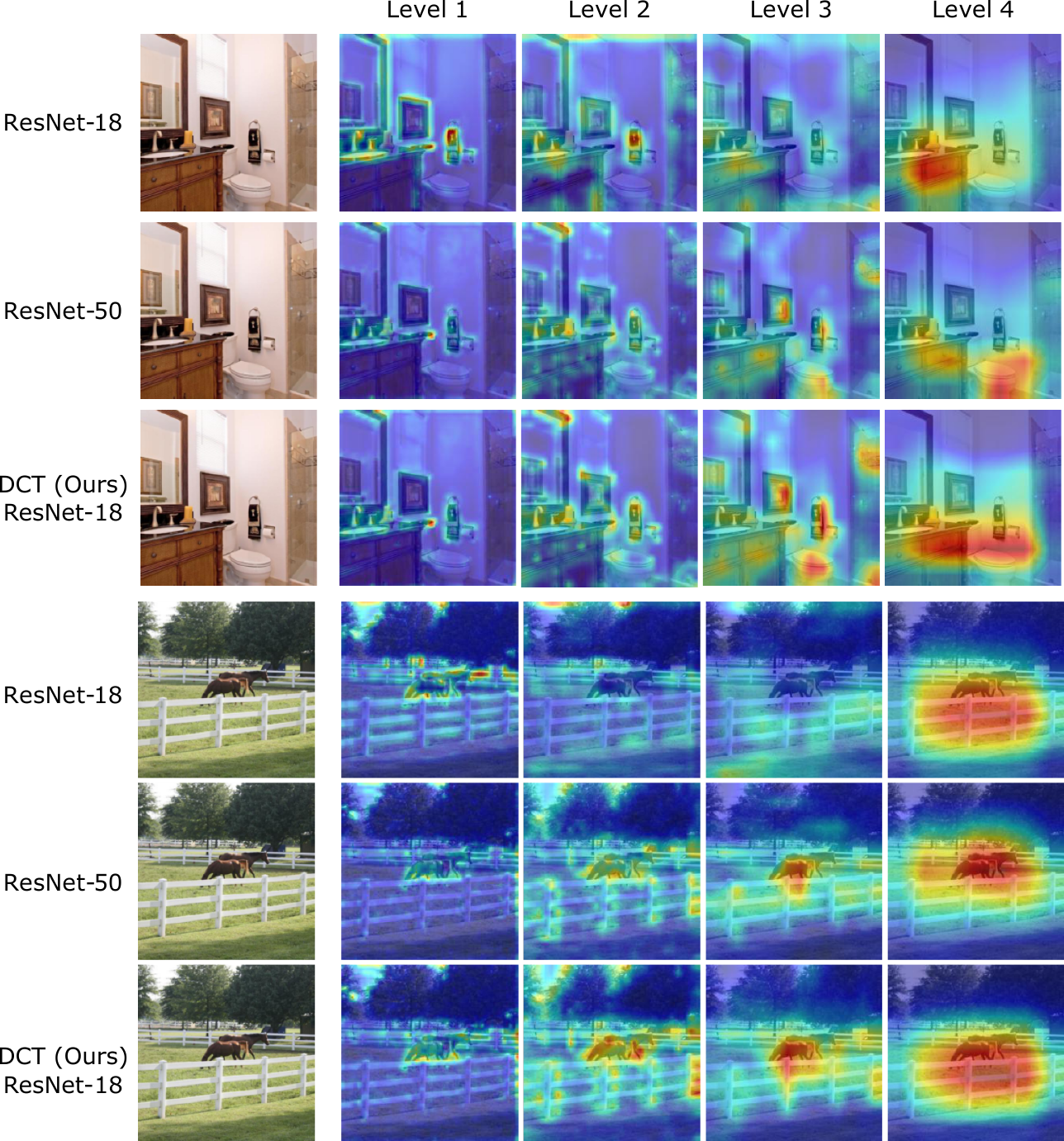}
    \caption{Obtained activation maps for the proposed method using ResNet-50 as teacher and ResNet-18 as student. Note how the proposed approach enables a ResNet-18 architecture to have similar activation maps to the ones obtained by a ResNet-50.}
    \label{fig:Comparison AMs}
\end{figure*}

\begin{figure*}[!t]
    \centering
    \includegraphics[width=0.8\textwidth, height=0.8\textheight, keepaspectratio]{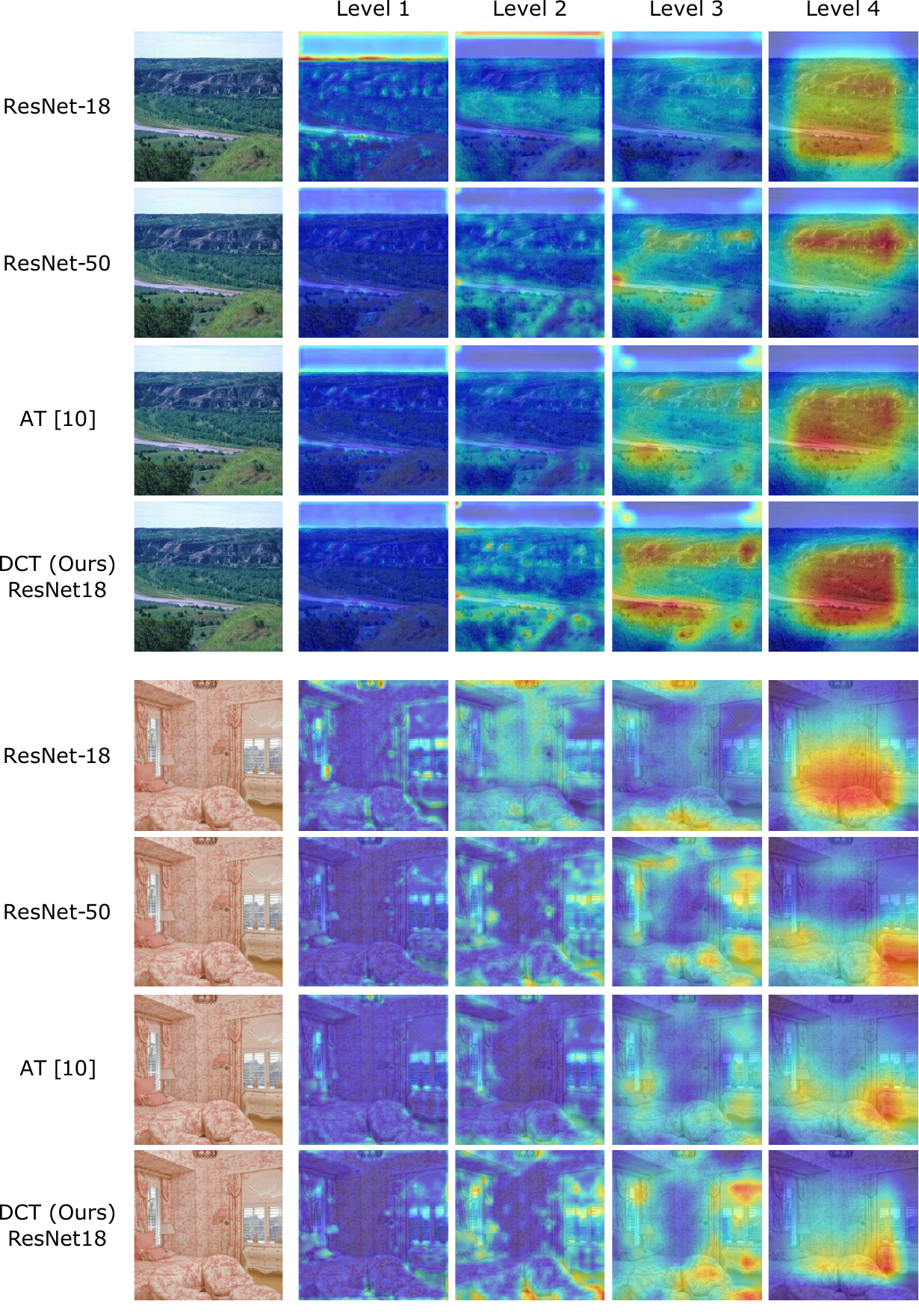}
    \caption{Obtained activation maps for the proposed method using ResNet-50 as teacher and ResNet-18 as student. AT \cite{komodakis2017paying} activation maps are also included for comparison. Note how the proposed approach enables a ResNet-18 architecture to have similar activation maps to the ones obtained by a ResNet-50. Note also how the matching is better than the one achieved by AT \cite{komodakis2017paying}.}
    \label{fig:Comparison AT}
\end{figure*}

\begin{table*}[t]
\scriptsize
\renewcommand{\arraystretch}{1.25}
\begin{centering}
\caption{Similarity between ResNet-50 activation maps, trained in ADE20K dataset, and the corresponding level's activation maps of several models. SSIM values close to 1 indicate identical maps and values close to 0 indicate no similarity.}
\resizebox{\textwidth}{!}{
\begin{tabular}{llccccccccc}
\hline 
\multirow{2}{*}{Method} & \multicolumn{5}{c}{Training Set} & \multicolumn{5}{c}{Validation Set}\tabularnewline
\cline{2-11} 
 & Level 1 & Level 2 & Level 3 & Level 4 & Average & Level 1 & Level 2 & Level 3 & Level 4 & Average\tabularnewline
\hline 
ResNet-18 & 0.46 & 0.32 & 0.39 & 0.72 & 0.48 & 0.47 & 0.32 & 0.40 & 0.71 & 0.47\tabularnewline
AT \cite{komodakis2017paying} & 0.66 & 0.73 & 0.76 & \textbf{0.90} & 0.76 & 0.67 & 0.74 & 0.77 & \textbf{0.83} & 0.75\tabularnewline
DCT (Ours) & \textbf{0.89} & \textbf{0.87} & \textbf{0.81} & 0.82 & \textbf{0.85} & \textbf{0.89} & \textbf{0.87} & \textbf{0.81} & 0.79 & \textbf{0.84}\tabularnewline
KD \cite{hinton2015distilling} & 0.48 & 0.55 & 0.42 & 0.78 & 0.56 & 0.48 & 0.56 & 0.43 & 0.73 & 0.56\tabularnewline
DCT (Ours) + KD & \textbf{0.90} & \textbf{0.88} & \textbf{0.82} & 0.87 & \textbf{0.87} & \textbf{0.90} & \textbf{0.88} & \textbf{0.83} & \textbf{0.83} & \textbf{0.86}\tabularnewline
\hline 
\end{tabular}}
\par\end{centering}
\label{tab:QuantitativeAM}
\end{table*}

Figures \ref{fig:ActivationMaps}, \ref{fig:Comparison AMs} and \ref{fig:Comparison AT} present qualitative results of the obtained activation maps by the proposed method. In addition, Figures \ref{fig:ActivationMaps} and \ref{fig:Comparison AT} include those obtained by AT \cite{komodakis2017paying} for comparison. Specifically,  Figure \ref{fig:ActivationMaps} shows how AT maps resemble teacher ones only in the wider and intense areas of activation, i.e., the \textit{bed} and the \textit{wardrobe} in Level 3, while the proposed approach yields more similar maps in all the image areas where the teacher is focused on, i.e., the \textit{bed}, and the \textit{wardrobe} but also the \textit{lamps}, the \textit{paintings} and even the \textit{book} on the table. This suggests that the proposed DCT-based metric achieves a better matching when activation patterns are diverse and spread throughout the image.

Table \ref{tab:QuantitativeAM} quantifies qualitative observations from Figures \ref{fig:ActivationMaps}, \ref{fig:Comparison AMs} and \ref{fig:Comparison AT} by repeating the presented experiment from Section \ref{subsubsec:CIFAR100}, i.e., computing the similarity between ResNet-50 (Teacher) and some model's activation maps for the whole set of training and validation samples in the ADE20K dataset using the SSIM.

Results in Table \ref{tab:QuantitativeAM} confirm the qualitative analysis presented in Figures \ref{fig:ActivationMaps}, \ref{fig:Comparison AMs} and \ref{fig:Comparison AT}: the similarity for levels $L={1..3}$, in both Training and Validation sets, increases when the proposed DCT-based loss is used. Level $L=4$ similarity is slightly better for AT, mainly because activation maps in this level tend to be image-centred, continuous, and mono-modal, which benefits the $\ell_2$ measure. Overall, the average similarity achieved by the proposed DCT method is \(11.84\%\) higher for the training set and \(12\%\) higher for the validation respect to AT. Finally, it is remarkable how similarity is even higher when the DCT+KD combination is used, which again indicates a high complementarity between both losses.

\section{Conclusions} \label{sec:Conclusions}
This paper proposes a novel approach to globally compare 2D structures or distributions by evaluating their similarity in the Discrete Cosine Transform domain. The proposed technique is the core of an Attention-based Knowledge Distillation method that aims to transfer knowledge from a teacher to a student model. Specifically, intermediate feature representations from the teacher and the student are used to obtain activation maps that are spatially matched using a DCT-based loss. The proposal is applied to the scene recognition task, where the attention of trained models is highly correlated with performance. The reported results show that the proposed approach outperforms the state-of-the-art Knowledge Distillation approaches via better comparing attention maps.

The presented results provide promising evidences that the use of 2D discrete linear transforms that efficiently capture 2D patterns might be helpful, not only for the Knowledge Distillation task, but also for other Computer Vision tasks where vectorial metrics, i.e. \(\ell_{2}\) metrics, are nowadays used by default.

\section*{Acknowledgments}
This study has been supported by the Spanish Government through the \textit{Formacion de Personal Investigador} (FPI) programm (PRE2018-084916 grant) from the TEC2017-88169-R MobiNetVideo project.

\bibliography{mybibfile}
\bibliographystyle{IEEEtran}

\end{document}